\documentclass[lettersize,journal]{IEEEtran}
\usepackage{amsmath,amssymb,amsfonts}
\usepackage{algorithmic}
\usepackage{algorithm}
\usepackage{array}
\usepackage{bm}
\usepackage{textcomp}
\usepackage{stfloats}
\usepackage{url}
\usepackage{verbatim}
\usepackage{graphicx}
\usepackage{textcomp}
\usepackage{xcolor}
\usepackage{dsfont}
\usepackage{bm,mathabx,amsthm}
\usepackage{subcaption}
\usepackage{graphicx}
\usepackage{booktabs}
\usepackage{romannum}
\newtheorem{theorem}{Theorem}

\newtheorem{remark}[theorem]{Remark}

\def\BibTeX{{\rm B\kern-.05em{\sc i\kern-.025em b}\kern-.08em
    T\kern-.1667em\lower.7ex\hbox{E}\kern-.125emX}}
\usepackage{balance}
\begin{document}
\title{Offline Diffusion Policy for Multi-User Delay-Constrained Scheduling}
\author{Zhuoran Li, Ruishuo Chen, Hai Zhong, Longbo Huang\textsuperscript{*},~\IEEEmembership{Senior Member,~IEEE}
\thanks{Manuscript created August, 2025; revised January, 2026; accepted April, 2026.  
Zhuoran Li, Ruishuo Chen, Hai Zhong and Longbo Huang are with Institute for Interdisciplinary Information Sciences (IIIS), Tsinghua University, Beijing 100084, China. \textsuperscript{*}Corresponding author: Longbo Huang (email: longbohuang@tsinghua.edu.cn). This work was supported by the National Natural Science Foundation
of China Grant 52494974.
}}

\markboth{IEEE Transactions on Mobile Computing, April~2026}%
{How to Use the IEEEtran \LaTeX \ Templates}

\maketitle

\begin{abstract}
Effective multi-user delay-constrained scheduling is crucial in various real-world applications, including embodied AI, instant messaging, live streaming, and data center management, where efficient resource allocation is required among users with diverse delay sensitivities. In these scenarios, schedulers must make real-time decisions to satisfy both delay and resource constraints without prior knowledge of system dynamics, which are often time-varying and challenging to estimate. {Current learning-based methods typically require online interactions with actual systems during the training stage. Therefore, these approaches are often difficult or impractical, as they can significantly degrade system performance and incur substantial service costs.} To address these challenges, we propose a novel offline reinforcement learning-based algorithm, named  
\underline{S}cheduling By \underline{O}ffline Learning with \underline{C}ritic Guidance and \underline{D}iffusion Model (SOCD), to learn efficient scheduling policies purely from pre-collected \emph{offline data}. SOCD innovatively employs a diffusion policy, complemented by a sampling-free critic network for policy guidance. By integrating the Lagrangian multiplier optimization into the offline reinforcement learning, SOCD efficiently trains high-quality constraint-aware policies exclusively from available datasets, eliminating the need for online interactions with the system.  
Experimental results demonstrate that SOCD is resilient to various system dynamics, including partially observable and large-scale environments, and delivers superior performance compared to existing methods.
\end{abstract}


\section{Introduction}
\IEEEPARstart{T}{he} rapid expansion of real-time interactive applications, such as instant messaging \cite{xiao2007understanding}, live streaming \cite{pires2015youtube}, embodied AI \cite{embodied_survey}, and data center management \cite{dayarathna2016energy}, has underscored the necessity for effective multi-user delay-constrained scheduling. Such scenarios often involve users with diverse latency sensitivities, requiring efficient allocation of limited resources under dynamic conditions \cite{wang2024appointment}. The quality of scheduling directly has a direct impact on user satisfaction and the economic viability of the system. For instance, \cite{kalankesh2020factors} shows that system response time is a predominant factor in user satisfaction, especially in mobile communication systems. Embodied AI systems, e.g., robotics \cite{robotic_survey}, intelligent manufacturing \cite{industry_survey}, and autonomous driving \cite{ad_survey}, require timely, resource-efficient sensor data transmission, which can be hindered by hardware, battery, or environmental constraints \cite{Ge_2023_ICCV}. Moreover, scheduling policies must respect resource budgets; for example, power consumption is a critical concern in datacenter maintenance \cite{chaoqiang2020a}, and resource-efficient scheduling can also promote low-carbon systems \cite{li2023low}.

{Satisfying rigorous delay bounds \cite{li2021delay} alongside strict resource constraints \cite{tang2020minimizing} necessitates real-time scheduling mechanisms capable of adapting to highly volatile environments. The surging user demands and network complexities inherent in contemporary large-scale systems, such as video streaming \cite{park2024cross}, the Internet of Things (IoT) \cite{ibrahim2024advancing}, and embodied AI \cite{industry_survey}, further highlight the critical need for scalable and highly adaptive scheduling policies. While conventional techniques like dynamic programming (DP) \cite{chen2018timely} and Lyapunov optimization \cite{liu2023learning} achieve optimality under precise models, they struggle in highly dynamic environments. Specifically, DP requires perfect prior knowledge of system dynamics, whereas Lyapunov targets long-term average stability rather than strict per-packet delay boundaries. Consequently, both are impractical in unmodeled, fast-changing scenarios. Although online learning approaches relax these strict modeling prerequisites, their reliance on continuous real-time trial-and-error often degrades Quality of Service (QoS) and remains prohibitive in safety-critical deployments.}

Machine learning, particularly deep reinforcement learning (DRL), has shown strong capability in handling environmental uncertainties such as non-stationarity and partial observability \cite{hu2024multi}. However, online DRL methods demand extensive real-time interactions with live systems, which can degrade service quality, incur high costs, or be infeasible in safety-critical settings \cite{wan2023scheduling}. In practice, such trial-and-error exploration is often costly and inherently risky, potentially disrupting services or causing instability. To avoid prolonged online exploration, we focus on \emph{offline RL algorithm} for scheduling to train effective policies purely from pre-collected data without environment interaction. 

{However, directly applying DRL techniques to the offline scheduling problem presents a significant challenge. Specifically, delay-constrained scheduling involves handling complex system dynamics and guaranteeing user latency. While existing methods such as SOLAR \cite{li2024offline} attempt to address this, they typically rely on simple policy representations with limited expressiveness, making them insufficient to capture the complex, multi-modal distributions inherent in optimal scheduling strategies. Consequently, offline RL algorithms often struggle to learn effective policies. This difficulty arises from the substantial policy inconsistency between the target policy and the unknown behavior policy that generates the offline data. Specifically, because the system prohibits online interactions, the algorithm cannot easily correct these inconsistencies. Therefore, training a high-quality policy becomes significantly more challenging when the offline data is of poor quality.}

{To address the unique challenges of offline multi-user delay-constrained scheduling under strict resource budgets, we propose \underline{S}cheduling by \underline{O}ffline Learning with \underline{C}ritic Guidance and \underline{D}iffusion Generation (SOCD). SOCD is specifically designed for queuing-network scheduling tasks that involve long-term temporal coupling, partial observability, and stringent constraint satisfaction.
Specifically, SOCD leverages offline reinforcement learning to learn directly from historical data, which avoids risky online exploration (e.g., RSD4 \cite{hu2024multi}). Furthermore, it integrates a diffusion-based policy network. In contrast to Gaussian-based baselines such as SOLAR \cite{li2024offline}, this network significantly enhances policy expressiveness and stability. To enforce both delay and resource constraints, a Lagrangian dual mechanism works alongside a sampling-free critic network that guides the policy updates.
This tailored integration of offline RL, diffusion modeling, and critic-guided constraint handling enables SOCD to achieve strong performance and robust constraint satisfaction in scenarios where existing methods struggle, representing, to the best of our knowledge, the first such design in delay-constrained network scheduling.}

{SOCD effectively addresses these challenges in the following ways.
First, it ensures compliance with resource and delay constraints by combining Lagrangian dual optimization for resource limits with delay-sensitive queues, all in a fully offline manner. Second, in contrast to traditional online RL methods—which can be unsafe and costly due to extensive real-time interactions—and purely model-based or Lyapunov approaches that may falter under complex, time-varying dynamics, SOCD learns entirely from offline data without disrupting system operations or incurring interaction costs. Finally, by formulating the problem as a Markov Decision Process (MDP), SOCD can adapt to diverse and dynamic system conditions.}

The contributions of this paper are threefold:
\begin{itemize}
    \item We propose a general offline learning formulation that effectively addresses multi-user delay-constrained scheduling problems with resource constraints. Our approach ensures compliance with both resource and delay constraints by utilizing the Lagrangian dual and delay-sensitive queues, all in a fully offline manner.
    \item {We introduce \underline{S}cheduling By \underline{O}ffline Learning with \underline{C}ritic Guidance and \underline{D}iffusion Generation (SOCD), a novel offline RL algorithm that integrates a critic-guided diffusion framework to efficiently train high-quality policies specifically for the resource and delay constrained scheduling problems. In contrast to conventional actor–critic methods, SOCD only trains its diffusion policy once, while the critic is adaptively updated without requiring target actions via diffusion policy, thereby enhancing both training efficiency and robustness.}

    \item We assess the performance of SOCD through comprehensive experiments conducted across diverse environments. Notably, SOCD demonstrates exceptional capabilities in  multi-hop networks, adapting to different arrival and service rates, scaling efficiently with a growing number of users, and operating effectively in partially observed environments, even without channel information. The results highlight SOCD's robustness to varying system dynamics and its consistent superiority over existing methods. 
\end{itemize}

The remainder of this paper is organized as follows. Section~\ref{sec:rw} reviews related work on delay-constrained scheduling, offline reinforcement learning and diffusion models. Section~\ref{sec:pf} presents the scheduling problem and its MDP formulation. Section~\ref{section:offline-rl} provides a detailed description of the SOCD algorithm. Section~\ref{sec:exp} discusses the experimental setup and presents the results. Finally, Section~\ref{sec:con} concludes the paper.

\section{Related Work}\label{sec:rw}
Numerous studies have addressed the scheduling problem through various approaches, including optimization-based methods and deep reinforcement learning. Additionally, recent advancements in offline reinforcement learning algorithms and diffusion models are reviewed in this section.

\subsection{Optimization-Based Scheduling}
\subsubsection{Stochastic Optimization}
{A key research direction for addressing scheduling problems involves the use of stochastic optimization techniques. This category encompasses four main methodologies: convex optimization, e.g., \cite{jung2020joint}, which focuses on network scheduling, dynamic programming (DP)-based control, e.g.,  \cite{chen2018timely}, which aims to achieve throughput-optimal scheduling, queueing theory-based methods, e.g.,  \cite{huang2015backpressure, mas2022queuing}, which concentrate on multi-queue systems, and Lyapunov  optimization, e.g., \cite{battiloro2022lyapunov}, used for network resource allocation and control. Although these methods are theoretically sound, they often encounter difficulties in explicitly addressing delay constraints and require precise knowledge of system dynamics, which is challenging to obtain in practical applications.}

\subsubsection{Combinatorial Optimization}
{Combinatorial optimization is another key class of methods for network scheduling. For instance, \cite{koosheshi2019optimization} employs fuzzy logic to optimize resource consumption with multiple mobile sinks in wireless sensor networks. More recently, graph neural networks (GNNs) have been applied to tackle scheduling problems, as demonstrated in \cite{li2023network}. Additionally, \cite{vesselinova2020learning} provides a comprehensive survey of applications in the domain of telecommunications. Despite their potential, these combinatorial optimization approaches often suffer from the curse of dimensionality, which makes it challenging to scale effectively to large-scale scenarios.}

\subsection{DRL-based Scheduling}
{Deep reinforcement learning (DRL) has garnered increasing attention in the field of scheduling, owing to its strong performance and significant potential for generalization. DRL has been applied in various scheduling contexts, including video streaming \cite{mao2017neural}, Multipath TCP control \cite{zhang2019reles}, and resource allocation \cite{meng2020power}. Despite the promise of DRL, these approaches often face challenges in enforcing average resource constraints and rely heavily on online interactions during training, which can be costly for many real-world systems. Offline reinforcement learning (Offline RL) algorithms are novel in that they require only a pre-collected offline dataset to train the policy without interacting with the environment \cite{levine2020offline}. The feature of disconnecting interactions with the environment is crucial for numerous application scenarios, such as wireless network optimization \cite{yang2024offline}. However, existing offline RL algorithms struggle to effectively address delay and resource constraints. Our proposed SOCD algorithm provides a robust, data-driven, end-to-end solution. It incorporates delay and resource constraints using delay-sensitive queues and the Lagrangian dual method, respectively. Additionally, it employs an offline learning paradigm, eliminating the need for online interactions.}

\subsection{Offline Reinforcement Learning and Diffusion Models}
{Distributional shift presents a significant challenge for offline reinforcement learning (RL), and various methods have been proposed to address this issue by employing conservatism to constrain the policy or Q-values through regularization techniques \cite{kumar2020conservative,song2020score}. Policy regularization ensures that the learned policy remains close to the behavior policy by incorporating policy regularizers. Examples include BRAC \cite{wu2019behavior}, BEAR \cite{kumar2019stabilizing}, BCQ \cite{fujimoto2019off}, TD3-BC \cite{fujimoto2021minimalist}, implicit updates \cite{nair2020awac}, and importance sampling approaches \cite{kostrikov2021offline}.  In contrast, critic regularization focuses on constraining Q-values to enhance stability, e.g., CQL \cite{kumar2020conservative}, IQL (Implicit Q-Learning) \cite{kostrikov2021offlines}, and TD3-CVAE \cite{rezaeifar2022offline}. Furthermore, developing a rigorous theoretical understanding of optimality and designing robust, well-founded algorithms remain fundamental challenges in the field of offline RL \cite{li2024towards}.}

{Diffusion models \cite{ho2020denoising}, a specialized class of generative models, have achieved remarkable success in various domains, particularly in generating images from text descriptions \cite{wang2024prolificdreamer}. Recent advancements have explored the theoretical foundations of diffusion models, such as their statistical underpinnings \cite{chen2023score}, and methods to accelerate sampling \cite{lu2022dpm}. Generative models have also been applied to policy modeling, including approaches like conditional VAE \cite{kingma2013auto}, Diffusers \cite{janner2022planning}, DQL \cite{wang2022diffusion}, SfBC \cite{chen2022offline}, and IDQL \cite{hansen2023idql}. }

{Our proposed SOCD represents a novel application of a critic-guided diffusion-based policy specifically designed for these scheduling tasks. However, distinct from generic offline RL algorithms (e.g., CQL \cite{kumar2020conservative} and IQL \cite{kostrikov2021offlines}) which focus primarily on mitigating distributional shift and often lack mechanisms to handle auxiliary constraints, SOCD integrates Lagrangian dual optimization directly into the offline framework. This allows strictly enforcing average resource and delay constraints without requiring online interaction. 
In contrast to standard offline RL and scheduling baselines (e.g., SOLAR \cite{li2024offline}) that typically rely on unimodal Gaussian policies, SOCD employs a diffusion-based policy. Therefore, it can successfully capture the complex, multi-modal action distributions inherent in multi-user queuing systems. Additionally, SOCD uniquely utilizes a sampling-free critic to decouple policy training from constraint updates. As a result, this design significantly enhances training efficiency compared to existing methods.
}

\section{Problem Formulation}\label{sec:pf}
This section presents our formulation for the offline learning-based scheduling problem. We begin by introducing the multi-user delay-constrained scheduling problem and its corresponding dual problem in Section~\ref{subsec:sm}. In Section~\ref{subsec:pomdp}, we present the MDP formulation for the dual problem. While our formulation builds on the foundations laid by previous works \cite{chen2018timely, hu2024multi, li2024offline}, our primary goal is to develop practical scheduling policies that can be implemented in real-world settings without the need for online interactions during training.

\subsection{The Multi-User Delay-Constrained Scheduling Problem}\label{subsec:sm}
\subsubsection{Single-Hop Setting}
\begin{figure}[htbp]
	\centering
\includegraphics[width=\columnwidth]{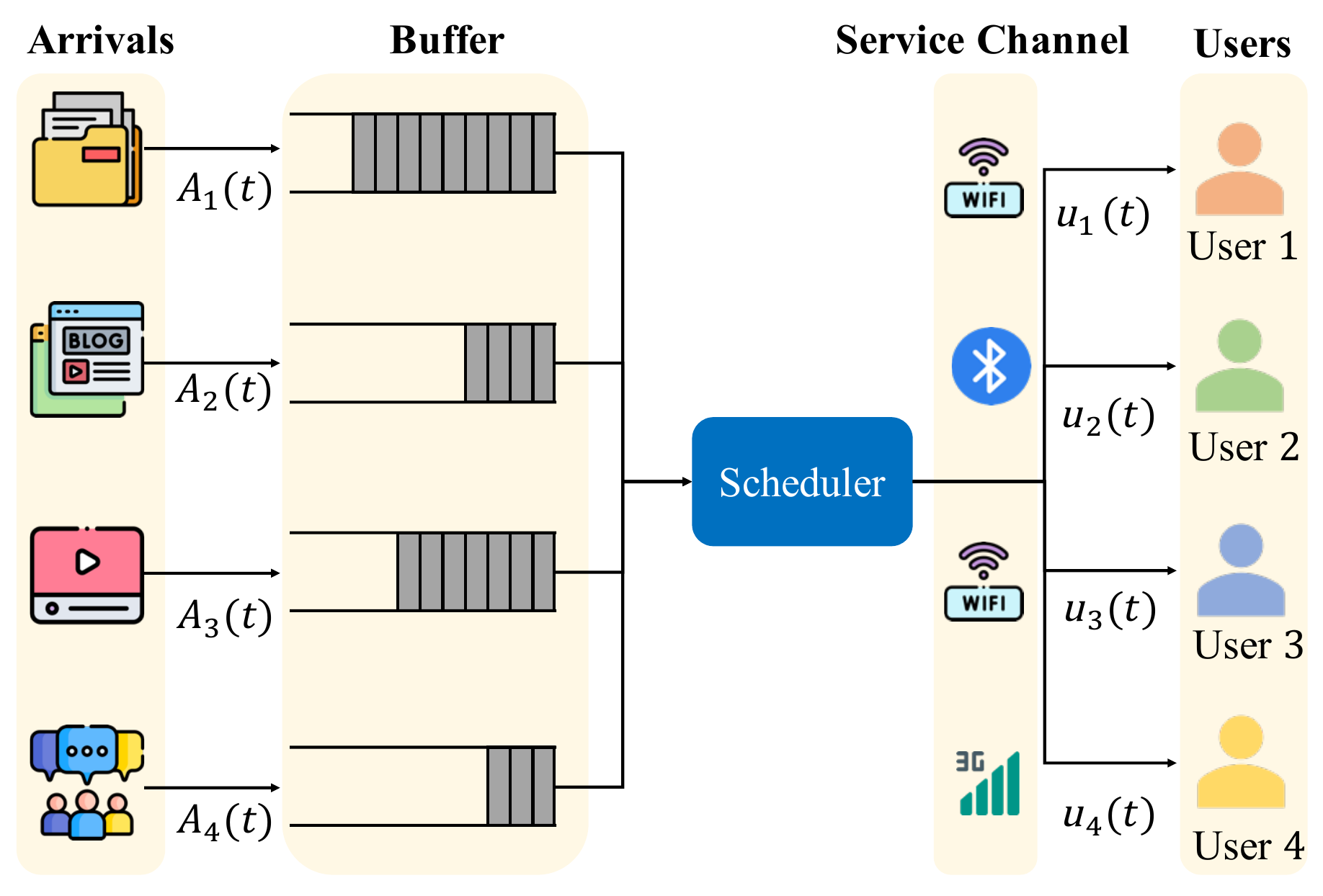}
	\caption{A four-user delay-constrained single-hop network.}
	\label{fig:sm}
\end{figure}
We consider the same scheduling problem as in \cite{singh2019throughput, hu2024multi,li2024offline}, using a four-user example shown in Figure~\ref{fig:sm}. Time is divided into discrete slots, \( t \in \{0, 1, 2, \dots\} \). The scheduler receives job arrivals, such as data packets in a network or parcels at a delivery station, at the beginning of each time slot. The number of job arrivals for user \( i \) at time slot \( t \) is denoted as \( A_i(t) \), and we define the arrival vector as \( \bm{A}(t) = [A_1(t), \dots, A_N(t)] \){, where $N$ refers to the number of users.} Each job for user \( i \) has a strict delay constraint \( \tau_i \), meaning that the job must be served within \( \tau_i \) slots of its arrival; otherwise, it will become outdated and be discarded.

\paragraph{The Buffer Model}
Jobs arriving at the system are initially stored in a buffer, which is modeled by a set of delay-sensitive queues. Specifically, the buffer consists of \( N \) separate delay-sensitive queues, one for each user, and each queue has infinite capacity. The state of queue \( i \) at time slot \( t \) is denoted by \( \bm{Q}_i(t) = [Q_i^{0}(t), Q_i^{1}(t), \dots, Q_i^{\tau_i}(t)] \), where $\tau_i$ is the maximum remaining time and \( Q_i^{\tau}(t) \) represents the number of jobs for user \( i \) that have \( \tau \) time slots remaining until expiration, for \( 0 \le \tau \le \tau_i \).

\paragraph{The Service Model}
At every time slot $t$, the scheduler makes decisions on the resources allocated to jobs in the buffer. For each user $i$, the decision is denoted as $\bm{v}_i(t)=[v_i^{0}(t), v_i^{1}(t), ..., v_i^{\tau_i}(t)]$, where $v_i^{\tau}(t) \in [0, v_{\max}]$ represents the amount of resource allocated to serve each job in queue $i$ with a deadline of $\tau$. Each scheduled job is then passed through a service channel, whose condition is random and represented by $c_i(t)$ for user $i$ at time slot $t$. The channel conditions at time slot $t$ are denoted as $\bm{c}(t)=[c_1(t), c_2(t), ..., c_N(t)]$. If a job is scheduled but fails in service, it remains in the buffer if it is not outdated. {The successful rate to serve each job is jointly determined by the channel condition $c_i(t)$ and $\bm{v}_i(t)$, i.e., $p_i(t) = P_i( \bm{v}_i(t), c_i(t))$.

\paragraph{The Scheduler Objective and Lagrange Dual}
For each user $i$, the number of jobs that are successfully served at time slot $t$ is denoted as $u_i(t)$. A known weight $\omega_i$ is assigned to each user, and the weighted instantaneous throughput is defined as $D(t) = \sum_{i=1}^N \omega_i u_i(t).$ The instantaneous resource consumption at time slot $t$ is denoted as $E(t) = \sum_{i=1}^N \bm{v}_i^\top(t) \bm{Q}_i(t), \label{eq:resource}$ and the average resource consumption is $\overline{E} = \lim_{T \to \infty} \frac{1}{T} \sum_{t=1}^T E(t)$. The objective of the scheduler is to maximize the weighted average throughput, defined as $\overline{D} = \lim_{T \to \infty} \frac{1}{T} \sum_{t=1}^T D(t)$, subject to the average resource consumption limit (corresponds to Equation (1) in \cite{hu2024multi} and Equation (3) in \cite{li2024offline}), i.e.,
\begin{align}\label{eq:p1}
	\mathcal{P}: \max_{\bm{v}_i(t): 1 \le i \le N, 1 \le t \le T} & \lim_{T \to \infty} \frac{1}{T} \sum_{t=1}^T \sum_{i=1}^N \omega_i u_i(t), \\
	\text{s.t.} & \lim_{T \to \infty} \frac{1}{T} \sum_{t=1}^T \sum_{i=1}^N \bm{v}_i^\top(t) \bm{Q}_i(t) \le E_0, \notag
\end{align}
where $E_0$ is the average resource consumption constraint.

We denote the optimal value of problem $\mathcal{P}$ by $\mathcal{T}^*$. Similar to Equation (2) in \cite{hu2024multi} and Equation (4) in \cite{li2024offline}, we define a Lagrangian function to handle the average resource budget constraint in problem $\mathcal{P}$, as follows:
\begin{equation}\label{eq:lg}	
	\mathcal{L}(\pi, \lambda) = \lim_{T \rightarrow \infty} \frac{1}{T} \sum_{t=1}^T \sum_{i=1}^N \left[\omega_i u_i(t) - \lambda \bm{v}_i^\top(t) \bm{Q}_i(t)\right] + \lambda E_0,
\end{equation}
where $\pi$ is the control policy and $\lambda$ is the Lagrange multiplier. The policy $\pi$ determines the decision value $\boldsymbol{v}_i(t)$ explicitly and influences the number of served jobs $u_i(t)$ implicitly. We denote $g(\lambda)$ as the Lagrange dual function for a given Lagrange multiplier $\lambda$: 
\begin{equation}\label{eq:lgd}
	g(\lambda) = \max_\pi \mathcal{L}(\pi, \lambda) = \mathcal{L}(\pi^*(\lambda), \lambda),
\end{equation}
where the maximizer is denoted as $\pi^*(\lambda)$.

\begin{remark}\label{rm:gd}
As shown in \cite{singh2019throughput, hu2024multi}, the optimal timely throughput, denoted by \( \mathcal{T}^* \), is equal to the optimal value of the dual problem, i.e., \( \mathcal{T}^* = \min_{\lambda \geq 0} g(\lambda) = g(\lambda^*) \), where \( \lambda^* \) is the optimal Lagrange multiplier. Moreover, the optimal policy \( \pi^*(\lambda^*) \) can be obtained by computing the dual function \( g(\lambda) \) for some \( \lambda \) and applying gradient descent to find the optimal \( \lambda^* \), as detailed in \cite{singh2019throughput, hu2024multi}. According to \cite{hu2024multi}, the derivative of the dual function \( g(\lambda) \) is given by \( g'(\lambda) = \frac{\partial \mathcal{L}(\pi^*(\lambda), \lambda)}{\partial \lambda} = E_0 - E_{\pi^*(\lambda)} \), where \( E_{\pi^*(\lambda)} = \lim_{T \to \infty} \frac{1}{T} \sum_{t=1}^{T} E_{\pi^*(\lambda)}(t) \) denotes the average resource consumption under the optimal policy \( \pi^*(\lambda) \).
\end{remark}

We emphasize that, although our formulation builds upon the work of \cite{singh2019throughput, hu2024multi}, our setting is significantly different. Our goal is to learn an efficient scheduling policy solely from pre-collected data, \emph{without} interacting with the system. Furthermore, our formulation ensures adherence to resource and delay constraints using the Lagrangian dual and delay-sensitive queues, all in a fully offline manner.

\subsubsection{Scalability in Multi-hop Setting}
\begin{figure*}[htbp]
    \centering
    
    \begin{subfigure}[c]{0.4\textwidth}
        \centering
        \includegraphics[width=\textwidth]{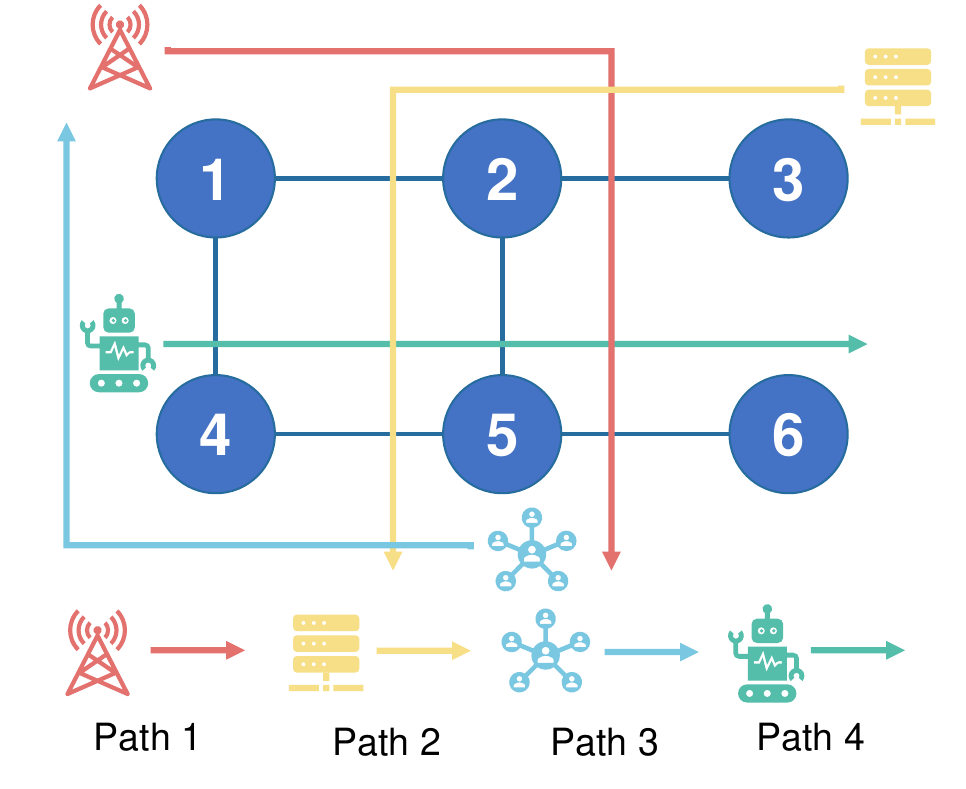}
        \caption{A multi-hop network.} 
        \label{fig:mh_topo}
    \end{subfigure}
    \hspace{0.5cm}
    \begin{subfigure}[c]{0.4\textwidth}
        \centering
        \includegraphics[width=\textwidth]{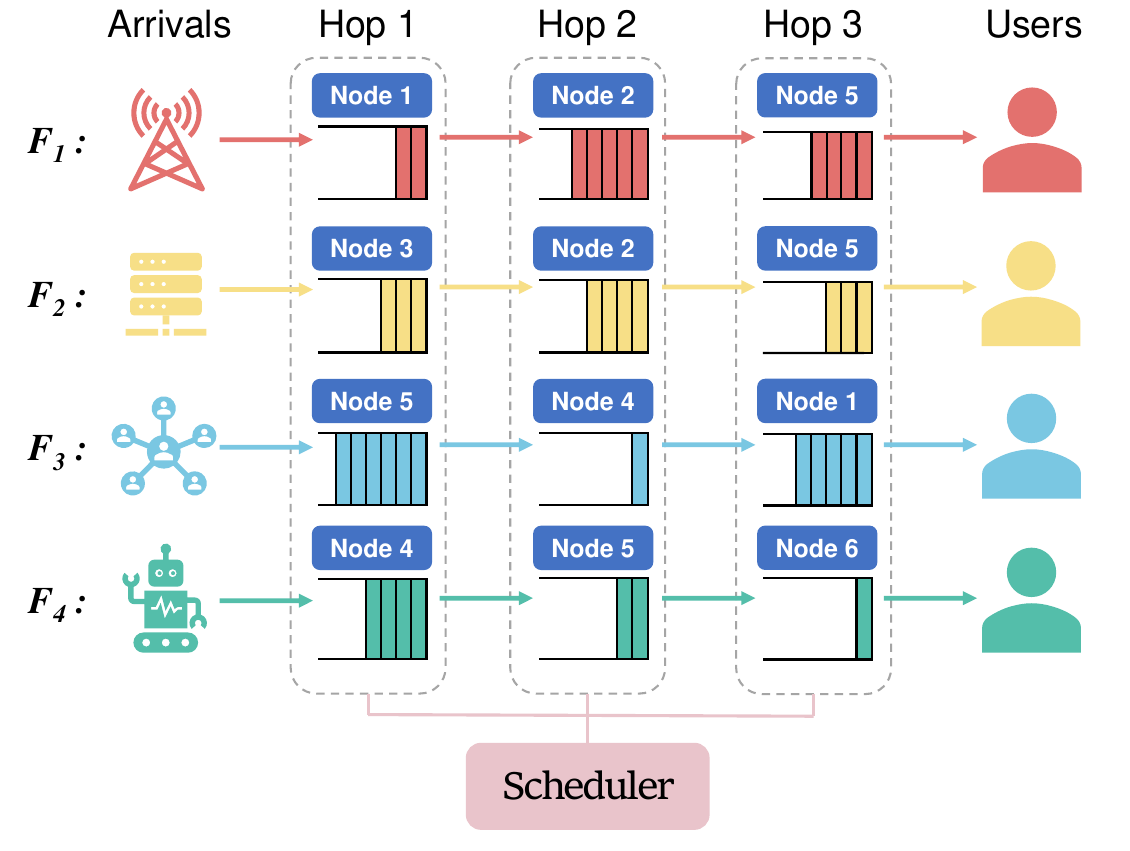}
        \caption{Four aligned flows.} 
        \label{fig:mh_flows}
    \end{subfigure}

    \caption{A four-user delay-constrained multi-hop network: (a) The multi-hop network with user flows, and (b) The buffer structure of the network.} 
    \label{fig:mh}
\end{figure*}

We consider a similar multi-hop formulation as in \cite{singh2019throughput, hu2024multi,li2024offline}. In a multi-hop network, each data flow spans multiple hops, requiring the scheduler to decide which flows to serve and how to allocate resources at each intermediate node. We consider a multi-hop network consisting of \( N \) flows (or users), denoted by \( \{F_1, F_2, \dots, F_N\} \), and \( K \) nodes, indexed as \( \{1, 2, \dots, K\} \). The length of the path for flow \( i \) is denoted by \( J_i \), and the longest path among all flows is given by \( J = \max_{1 \leq i \leq N} J_i \).

The routing of each flow \( i \) is encoded in a matrix \( \mathbf{H}^{(i)} = (h_{jk}^{(i)}) \in \mathbb{R}^{J \times K} \), where \( h_{jk}^{(i)} = 1 \) indicates that the \( j \)-th hop of flow \( i \) corresponds to node \( k \), and \( h_{jk}^{(i)} = 0 \) otherwise. At time \( t \), the number of jobs arriving for flow \( i \) is expressed as \( A_i(t) \), with each job subject to a strict delay requirement \( \tau_i \). Furthermore, each node must adhere to an average resource constraint \( E_0^{(k)} \).

As an example, Figure~\ref{fig:mh} illustrates a multi-hop network with four distinct flows. These flows traverse the network along the following routes: \( F_1 = \{1 \rightarrow 2 \rightarrow 5\} \), \( F_2 = \{3 \rightarrow 2 \rightarrow 5\} \), \( F_3 = \{5 \rightarrow 4 \rightarrow 1\} \), and \( F_4 = \{4 \rightarrow 5 \rightarrow 6\} \).

The multi-hop scheduling problem can be viewed as an extension of multiple single-hop scheduling problems, tailored to account for the unique characteristics of multi-hop scenarios. These characteristics include the buffer model, scheduling and service mechanisms, and the overarching system objectives specific to multi-hop networks.

\paragraph{Aggregated Buffer Model}  
In multi-hop networks, buffers are represented as delay-sensitive queues, similar to those in single-hop networks. However, due to the requirement for jobs to traverse multiple hops to reach their destinations, the buffer model must account for the flow of jobs across the entire network. To effectively describe the system state, we introduce an aggregated buffer model that organizes flows based on their originating nodes, as illustrated in Figure~\ref{fig:mh}.

For a flow $i$ with a path length of $J_i$, the buffer state at the $j$-th hop ($1 \leq j \leq J_i$) is denoted by $\bm{Q}_i^{(j)}(t) = [Q_i^{(j,0)}(t), Q_i^{(j,1)}(t), \dots, Q_i^{(j,\tau_i-j+1)}(t)]$. Here, $Q_i^{(j,\tau)}(t)$ represents the number of jobs in flow $i$ at the $j$-th hop with $\tau$ time slots remaining until expiration, where $0 \leq \tau \leq \tau_i-j+1$. For hops beyond the path length of flow $i$ ($J_i < j \leq J$), the buffer state is set to $\bm{Q}_i^{(j)}(t) = \bm{0}$, indicating the absence of flow $i$ in these hops.  

This formulation leads to the definition of $J$ aggregated buffers, represented as $\mathbf{Q}^{(1)}(t), \mathbf{Q}^{(2)}(t), \dots, \mathbf{Q}^{(J)}(t)$. Each aggregated buffer $\mathbf{Q}^{(j)}(t)$ consolidates the states of all flows at the $j$-th hop, such that $\mathbf{Q}^{(j)}(t) = [\bm{Q}_1^{(j)}(t), \bm{Q}_2^{(j)}(t), \dots, \bm{Q}_N^{(j)}(t)]$ for $1 \leq j \leq J$.

\paragraph{Multi-hop Scheduling and Service Model} 
At each time slot \( t \), the scheduler is responsible for deciding which flows to prioritize and determining the resource allocation for each node. The resources allocated to flow \( i \) are represented by \( \mathbf{v}_i(t) = [\bm{v}_i^{(1)}(t), \bm{v}_i^{(2)}(t), \dots, \bm{v}_i^{(J_i)}(t)] \), where \( \bm{v}_i^{(j)}(t) \) specifies the resources assigned to the \( j \)-th hop of flow \( i \), for \( 1 \leq j \leq J_i \). Consequently, the complete scheduling decision for all flows at time \( t \) is denoted as \(  \mathbf{V}(t) = [\mathbf{v}_1(t), \mathbf{v}_2(t), \dots, \mathbf{v}_N(t)] \).

Each scheduled job must traverse the service channels along its designated flow path. The service channel condition between nodes $i$ and $j$ at time $t$ is denoted by $c_{ij}(t)$. For a job belonging to flow $i$, given an allocated resource $v$ and channel condition $c$, the probability of successful service is $P_i(v, c)$. 
The instantaneous resource utilization at node $k$ during time slot $t$ is defined as $E^{(k)}(t) = \sum_{i=1}^N \sum_{j=1}^{J_i} h_{jk}^{(i)} {\bm{v}_i^{(j)}}^\top(t) \bm{Q}_i^{(j)}(t),$ where $h_{jk}^{(i)}$ represents the association of flow $i$'s $j$-th hop with node $k$, and $\bm{Q}_i^{(j)}(t)$ denotes the buffer state. The average resource utilization for node $k$ is given by $\overline{E}^{(k)} = \lim_{T \to \infty} \frac{1}{T} \sum_{t=1}^T E^{(k)}(t).$

\paragraph{Multi-hop System Objective and Lagrange Dual}  
The number of jobs successfully served for flow \( i \) at time slot \( t \) is denoted by \( u_i(t) \). Each flow is assigned a weight \( \omega_i \), and the weighted instantaneous throughput is given by \( D(t) = \sum_{i=1}^N \omega_i u_i(t) \). The scheduler's primary objective is to maximize the weighted average throughput, defined as \( \overline{D} = \lim_{T \to \infty} \frac{1}{T} \sum_{t=1}^T D(t) \), while ensuring compliance with the average resource consumption constraints at each node (similar to Equation (4) in \cite{hu2024multi}).

\vspace{-.2cm}  
\begin{eqnarray}\label{eq:p1_m}  
	\mathcal{P}_m:  
	\max_{\bm{v}_i(t): 1 \le i \le N, 1 \le t \le T} && \lim_{T \to \infty} \frac{1}{T} \sum_{t=1}^T \sum_{i=1}^N \omega_i u_i(t), \\  
	\text{s.t.} && \lim_{T \to \infty} \frac{1}{T} \sum_{t=1}^T \sum_{i=1}^N \sum_{j=1}^{J_i} h_{jk}^{(i)} {\bm{v}_i^{(j)}}^\top(t)\cdot \notag \\  
	&& \bm{Q}_i^{(j)}(t) \le E_0^{(k)}, \quad \forall 1 \le k \le K, \notag  
\end{eqnarray}  
where $E_0^{(k)}$ represents the average resource constraint at node $k$.  
The Lagrangian dual function for the multi-hop optimization problem $\mathcal{P}_m$ is expressed as: 
\begin{align}
\mathcal{L}_m(\pi, \bm{\lambda}) = &\lim_{T \to \infty} \frac{1}{T} \sum_{t=1}^T \sum_{i=1}^N \Big[ \omega_i u_i(t) \label{eq:lgm} \\ 
& - \sum_{k=1}^K \lambda_k \sum_{j=1}^{J_i} h_{jk}^{(i)} {\bm{v}_i^{(j)}}^\top(t) \bm{Q}_i^{(j)}(t) \Big] + \sum_{k=1}^K \lambda_k E_0^{(k)}, \notag
\end{align}  
where $\bm{\lambda} = [\lambda_1, \lambda_2, \dots, \lambda_K]$ is the vector of Lagrange multipliers.  

The Lagrange dual function, given a fixed multiplier $\bm{\lambda}$, is defined as:  
\begin{equation}
g_m(\bm{\lambda}) = \max_\pi \mathcal{L}_m(\pi, \bm{\lambda}) = \mathcal{L}_m(\pi_m^*(\bm{\lambda}), \bm{\lambda}),
\end{equation}  
where $\pi_m^*(\bm{\lambda})$ represents the policy that maximizes the Lagrangian function.  
To derive the optimal policy $\pi_m^*(\bm{\lambda}^*)$, the dual function $g_m(\bm{\lambda})$ is optimized using gradient descent to identify the optimal multiplier $\bm{\lambda}^*$.

\subsubsection{Offline Learning}
Although our formulation bears similarities to those in \cite{hu2024multi, li2024offline}, our offline learning-based scheduling setting introduces unique challenges. Specifically, the algorithm cannot interact with the environment during training and must rely entirely on a pre-collected dataset \( \mathcal{D} \), which is generated by an unknown behavior policy that governs the data collection process.  
Importantly, the quality of \( \mathcal{D} \) may be limited or suboptimal, as it reflects the biases and imperfections of the behavior policy. The objective, therefore, is to train an effective policy based solely on the dataset, without any online interaction with the system.

\subsection{The MDP Formulation}\label{subsec:pomdp}
We present the Markov Decision Process (MDP) formulation for our scheduling problem, which allows  SOCD to determine the optimal policy $\pi(\lambda)$. 
Specifically, an MDP is defined by the tuple $\mathcal{M} = \langle \mathcal{S}, \mathcal{A}, r, \mathbb{P}, \gamma \rangle$, where $\mathcal{S}$ represents the state space, $\mathcal{A}$ is the action space, $r$ is the reward function, $\mathbb{P}$ is the transition matrix, and $\gamma$ is the discount factor. The overall system state $\bm{s}_t$ includes $\bm{A}(t)$, $\bm{Q}_1(t)$, ..., $\bm{Q}_N(t)$, $\bm{c}(t)$, and possibly other relevant information related to the underlying system dynamics. At each time slot $t$, the action is represented by $\bm{a}_t = [\bm{v}_1(t), ..., \bm{v}_N(t)]$. 
Similar to \cite{li2024offline}, to take the resource constraint into consideration, 
 we set the reward of the MDP to be 
\begin{equation}
	r_t = D(t) - \lambda E(t), \label{eq:reward}
\end{equation}
where $D(t)$ is the instantaneous weighted throughput, and $E(t)$ is the resource consumption weighted by $\lambda$. 

The objective of the optimal agent is to learn an optimal policy \( \pi(\cdot; \bm{\theta}) \), parameterized by \( \bm{\theta} \), which maps the state to the action space in order to maximize the expected rewards:
\begin{equation}
    \begin{aligned}
        J(\pi(\cdot; \bm{\theta})) = \mathbb{E}\left[\sum_{t=0}^{T} \gamma^t r(s_t, a_t) \mid s_0, a_0, \pi(\cdot; \bm{\theta})\right].
    \end{aligned}
    \label{eq:ltr}
\end{equation}

\begin{remark}
By considering the above MDP formulation, our objective is to maximize the Lagrange function under a fixed $\lambda$, i.e., to obtain the Lagrange dual function in Equation~\eqref{eq:lgd}. Specifically, the reward setting in Equation~\eqref{eq:reward} implies that when $\gamma = 1$, the cumulative discounted reward becomes $R = \sum_{t=1}^T \gamma^t r_t = T {\mathcal{L}}(\pi, \lambda) - \lambda T E_0$. Therefore, an algorithm that maximizes the expected rewards $J(\pi(\cdot; \phi))$ also maximizes the Lagrangian function, which is the objective of SOCD as detailed in Section~\ref{section:offline-rl}. Furthermore, we emphasize that our goal is to learn an efficient scheduling policy based purely on pre-collected data {without} interacting with the system.
\end{remark}

\section{SOCD: Offline RL-based Scheduling}\label{section:offline-rl}
{This section presents SOCD, our offline RL-based scheduling algorithm, and highlights its key innovations. SOCD is specifically designed to address multi-user scheduling problems under stringent delay and resource constraints in a fully offline setting. The framework consists of two core components: a critic-guided diffusion policy for offline reinforcement learning (Section~\ref{subsec:olar}) and Lagrange multiplier optimization for constraint satisfaction (Section~\ref{subsec:lg}). 

{Instead of conventional actor–critic methods, SOCD trains the diffusion policy in a single unified phase, eliminating the need for separate actor updates. Meanwhile, the critic is adaptively updated without relying on target actions sampled from the diffusion policy. This design significantly enhances both training efficiency and robustness, especially in fully offline settings.}
The overall structure of SOCD is illustrated in Figure~\ref{fig:overview}, demonstrating how these components work together to enable high-quality policy learning under challenging scheduling constraints.}

\begin{figure*}[htbp]
\centering
\includegraphics[width=0.8\linewidth]{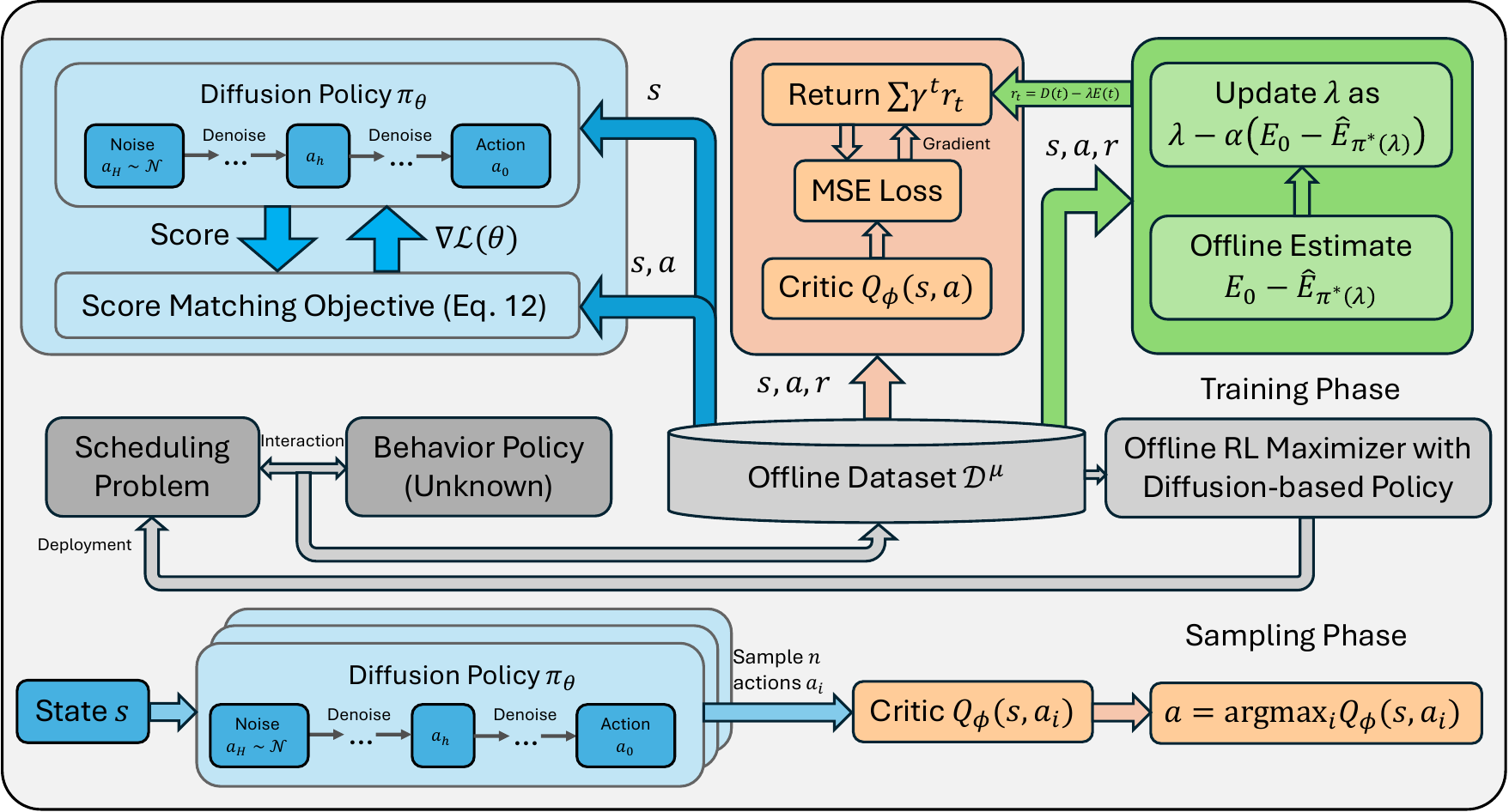}
\caption{{The SOCD algorithm: Operating solely in the offline phase, it does not require online interactions or prior knowledge of system dynamics. This makes it particularly well-suited for scenarios where online system interactions are either impractical or infeasible.}}
\label{fig:overview}
\end{figure*}

\subsection{Offline RL Maximizer with Diffusion-based Policy}\label{subsec:olar}
In this subsection, we introduce our offline reinforcement learning (RL) algorithm, which consists of two key components: a diffusion model for behavior cloning and a critic model for learning the state-action value function (Q-value). The primary objective is to maximize the Lagrangian function in Equation~\eqref{eq:lgd} for a fixed Lagrange multiplier $\lambda$, without requiring online interactions.

We begin by considering the general offline RL setting, where the Markov Decision Process (MDP) is defined as $\langle \mathcal{S}, \mathcal{A}, r, \mathbb{P}, \gamma \rangle$, and the offline dataset $\mathcal{D}^\mu = \left\{\mathcal{T}_j = (\bm{s}^j_1, \bm{a}^j_1, \dots, \bm{s}^j_t, \bm{a}^j_t, \dots, \bm{s}^j_T, \bm{a}^j_T)\right\}_{j=1}^{J}$ is collected by a behavior policy $\mu(\cdot|\bm{s})$. 
We define the state-action value function (i.e., the critic model) as $Q(\bm{s}, \bm{a})$\footnote{We clarify that $Q(\bm{s},\bm{a})$ is a state-action function in reinforcement learning, which is different from the buffer size of user $i$ in time $t$ notated as $\bm{Q}_i(t)$.} and the policy as $\pi(\cdot|\bm{s})$.

To ensure the learned policy remains close to the behavior policy while maximizing the expected value of the state-action functions, prior work \cite{peng2019advantage, nair2020awac} has explicitly constrained the policy $\pi$ using the following formulation:
\begin{align}
    \mathop{\mathrm{arg \ max}}_{\pi} \quad &\int_\mathcal{S} \rho_\mu(\bm{s}) \int_\mathcal{A} \pi(\bm{a} | \bm{s}) Q(\bm{s}, \bm{a}) \mathrm{d}\bm{a} \mathrm{d}\bm{s} \nonumber \\ 
    - \frac{1}{\alpha}&\int_\mathcal{S} \rho_\mu(\bm{s}) D_{\mathrm{KL}} \left(\pi(\cdot |\bm{s}) ||  \mu(\cdot |\bm{s}) \right) \mathrm{d}\bm{s}.
    \label{eq:rl_main}
\end{align}
Here, $\rho_\pi(\bm{s}) = \sum_{n=0}^\infty \gamma^n p_\pi(\bm{s}_n = \bm{s})$ denotes the discounted state visitation frequencies induced by policy $\pi$, with $p_\pi(\bm{s}_n=\bm{s})$ representing the probability of the event $\{\bm{s}_n = \bm{s}\}$ under policy $\pi$. The optimal policy $\pi^*$ of Equation~\eqref{eq:rl_main} can be derived using the method of Lagrange multipliers \cite{peng2019advantage, nair2020awac}: 
\begin{align}
    \pi^*(\bm{a}|\bm{s}) &= \frac{1}{Z(\bm{s})} \mu(\bm{a}|\bm{s}) \mathrm{exp}\left(\alpha Q(\bm{s}, \bm{a}) \right),
\label{eq:pi_optimal}
\end{align}
where $Z(\bm{s})$ is the partition function and $\alpha$ is he
temperature coefficient.  

A key observation based on Equation~\eqref{eq:pi_optimal} is that  $\pi^*$ can be decoupled into two components: (\romannum{1}) a behavior cloning (BC) model $\mu_{\bm{\theta}}(\cdot|\bm{s})$ trained on $\mathcal{D}^\mu$ parameterized by $\bm{\theta}$, which is irrelevant to the Lagrange multiplier $\lambda$, and (\romannum{2}) a critic model $Q_{\bm{\phi}}(\bm{s}, \bm{a})$ parameterized by $\bm{\phi}$, which is influenced by the Lagrange multiplier $\lambda$. 

\subsubsection{Diffusion-based Behavior Cloning}
It is critical for the BC model to express multi-modal action distributions, as previous studies \cite{wang2022diffusion} have pointed out that uni-modal policies are prone to getting stuck in suboptimalities. Moreover, offline RL methods often suffer from value overestimation due to the distributional shift between the dataset and the learned policy \cite{kumar2020conservative}, underscoring the importance of maintaining high fidelity in the BC model. To address these challenges, we propose cloning the behavior policy using diffusion models, which have demonstrated significant success in modeling complex and diverse distributions \cite{Qin2023Class-Balancing}.

We adopt the score-based structure introduced in \cite{song2020score}. To be specific, the forward process of the diffusion model gradually adds Gaussian noise to the action distribution conditional on the state given by the behavior policy, i.e., $p_0 (\bm{a}_0|\bm{s})=\mu (\bm{a}_0|\bm{s})$, such that the conditional transition function is given by $p_{t0} (\bm{a}_t|\bm{a}_0, \bm{s})=\mathcal{N}(\bm{a}_t;\alpha_t\bm{a}_0,\sigma_t^2\bm{I})$ for some $\alpha_t,\sigma_t>0$ and with a sufficiently small $\alpha_T$,  $p_{T0} (\bm{a}_T|\bm{a}_0, \bm{s})\approx\mathcal{N}(\bm{a}_T;0,\bm{I})$.\footnote{We omit the subscript $\mu$ from $p_\mu$ for the sake of simplicity and clarity. In this and the following paragraph, we use $\bm{a}_t$ to represent the action at time $t$ in both the forward and reverse process of the diffusion model instead of the trajectories, with a certain reuse of notation for clarity.} 
This corresponds to a stochastic differential equation (SDE) $\mathrm{d}\bm{a}_t = f(\bm{a}_t, t) \mathrm{d}t + g(t) \mathrm{d} \bm{w}_t$, where $\bm{w}_t$ is a standard Wiener process and $f(\cdot,t)$, $g(t)$ are called drift coefficient and diffusion coefficient, respectively \cite{song2020score}. Following the Variance Preserving (VP) SDE setting as introduced in \cite{song2020score}, we choose $\alpha_t = e^{-\frac{1}{2} \int_0^t \omega(s) \mathrm{d} s}$ and $\sigma_t^2 = 1-e^{-\int_0^t \omega(s) \mathrm{d} s}$, where $\omega(t)=\left(\omega_{\max }-\omega_{\min }\right)t+\omega_{\min}$ such that $f(\boldsymbol{a}_t,t) = \boldsymbol{a}_t\frac{{\rm{d}}\log \alpha_t}{{\rm{d}}t}$, $g^2(t) = \frac{{\rm{d}}\sigma_t^2}{{\rm{d}}t} - 2\frac{{\rm{d}}\log \alpha_t}{{\rm{d}}t}\sigma_t^2$.

Based on the findings in \cite{song2020score,kingma2021variational}, a corresponding reverse stochastic differential equation (SDE) process exists, enabling the transition from $T$ to $0$ that guarantees that the joint distribution $p_{0:T}(\bm{a}_0,\bm{a}_1,...,\bm{a}_T)$ are consistent with the forward SDE process. This process is detailed in Equation (2.4) of \cite{lu2022dpm} that considers the state $\boldsymbol{s}$ as conditioning variables:
\begin{equation} {\rm{d}}\boldsymbol{a}_{t} = \big[f(\bm{a}_t, t) - g^2(t)\underbrace{\nabla_{\bm{a}_t}p_{t}(\bm{a}_t|\boldsymbol{s})}_{\text{Neural Network } \boldsymbol{s}_{\boldsymbol{\theta}}}\big]{\rm{d}}t + g(t){\rm{d}}\overline{\boldsymbol{w}}_t,\label{eqreq:invsde} \end{equation}
where $p_{t}(\bm{a}_t|\boldsymbol{s})=\int_{\boldsymbol{a_0}}p_{t0}(\bm{a}_t|\bm{a}_0,\boldsymbol{s})\mu(\boldsymbol{a}_0|\boldsymbol{s}){\rm{d}}\bm{a}_0$ is the state-conditioned marginal distribution of $\boldsymbol{a}_t$ including the unknown data distribution $\mu$, and $\overline{\boldsymbol{w}}_t$ represents a standard Wiener process when time flows backwards from $T$ to $0$.

To fully characterize the reverse SDE process described in Equation~\eqref{eqreq:invsde}, it is essential to obtain the conditional \emph{score function}, defined as $\nabla_{\bm{a}_t}p_{t}(\bm{a}_t|\boldsymbol{s})$, at each $t$. The score-based model is to learn the conditional score model $\mathbf{s}_{\bm{\theta}}(\bm{a}_t, \bm{s}, t)$ that estimates the conditional score function of the behavior action distribution by minimizing the score-matching objective \cite{song2020score}:
\begin{equation}
\mathcal{L}(\bm{\theta}) = 
\mathbb{E}_{(\bm{s}, \bm{a}) \sim D^\mu,\bm{\epsilon}, t}[\| \sigma_t \mathbf{s}_{\bm{\theta}}(\alpha_t\bm{a}+\sigma_t\bm{\epsilon}, \bm{s}, t) + \bm{\epsilon} \|_2^2],
\label{eq:train_dif}
\end{equation}
where $\bm{\epsilon}\sim \mathcal{N}(0,\bm{I})$, $t\sim\mathcal{U}(0,T)$. When sampling, the reverse process of the score-based model solves a diffusion ODE from time $T$ to time $0$ to obtain $\bm{a}_0 \sim \mu_{\bm{\theta}}(\cdot|\bm{s})$:
\begin{equation}
\mathrm{d} \bm{a}_t = \bigg[f(\bm{a}_t, t) - \frac{1}{2}g^2(t) \mathbf{s}_{\bm{\theta}}(\bm{a}_t,\boldsymbol{s},t)\bigg] \mathrm{d}t.
\label{eq:solve_ode}
\end{equation}
with $\bm{a}_T$ sampled from $\mathcal{N}(0,\bm{I})$. In this manner, we can effectively perform behavior cloning using a diffusion policy.

It is important to note that training the diffusion policy is independent of the Lagrange multiplier $\lambda$, which means that the policy \emph{only needs to be trained once} compared with other offline RL-based scheduling algorithm, e.g., SOLAR \cite{li2024offline} needs multiple iterations of the policy for different Lagrange multiplier $\lambda$. This significantly enhances the model's reusability and reduces training time.

\subsubsection{Sampling-free Critic Learning}
The objective of training the state-action value function \( Q_{\bm{\phi}}(\bm{s}, \bm{a}) \) is to minimize the temporal difference (TD) error under the Bellman operator, which is expressed as:
\begin{equation}
\label{eq:sorarcritic}
    \begin{aligned}
        \mathcal{L}(\bm{\phi}) = \
        &\mathbb{E}_{(\bm{s}, \bm{a}, \bm{r}, \bm{s}') \sim \mathcal{D}^\mu, \bm{a}' \sim \pi(\cdot | \bm{s}')} \\
        & \left[(\gamma \min_{k=1,2} Q_{\overline{\bm{\phi}}}^k (\bm{s}', \bm{a}') + \bm{r} - Q_{\bm{\phi}}(\bm{s}, \bm{a}))^2 \right].
    \end{aligned}
\end{equation}

{However, calculating the TD loss requires generating new, and potentially unseen, actions $\bm{a}'$. This process is computationally inefficient, particularly when the system relies on multiple steps of denoising under diffusion sampling. Furthermore, generating out-of-sample actions introduces potential extrapolation errors \cite{fujimoto2019off}. Consequently, these errors can severely degrade the quality of the value estimates.}

To address these challenges, we propose a sampling-free approach for training the critic by directly using the discounted return as the target. Specifically, for each trajectory $\mathcal{T} = (\bm{s}_1, \bm{a}_1, \bm{r}_1, \dots, \bm{s}_t, \bm{a}_t, \bm{r}_t, \dots, \bm{s}_T, \bm{a}_T, \bm{r}_T)$ with rewards calculated via Equation~\eqref{eq:reward}, which is randomly sampled from the offline dataset $\mathcal{D}^\mu$, we define the critic's training objective as:
\begin{equation}
\label{eq:criticloss}
    \begin{aligned}
        \mathcal{L}(\bm{\phi}) =\mathbb{E}_{t \sim \mathcal{U}(1,T),\mathcal{T}\sim\mathcal{D}^{\mu}}\left[(\min_{k=1,2} Q_{\bm{\phi}}^k (\bm{s}_t, \bm{a}_t) -
        \sum_{i=t}^{T} \gamma^{i-t} \bm{r}_i )^2\right].
    \end{aligned}
\end{equation}
{By incorporating the double Q-learning technique \cite{hasselt2010double}, our method computes the TD error in a {sampling-free} manner—unlike conventional approaches, it does not require sampling target actions from the policy network. This avoids the multi-step denoising process inherent to diffusion-based policy sampling, whose performance is sensitive to the choice of $\lambda$. As a result, the critic training becomes significantly more efficient and scalable, while retaining stable loss convergence and mitigating overestimation bias in Q-value estimation, leading to consistently strong evaluation performance.}

\subsubsection{Selecting From Samples}
We can generate the final output action from the trained BC model (diffusion policy) under the guidance of the trained critic model. The action-generation procedure is shown in Algorithm~\ref{alg:socdgene}.

According to Equation~\eqref{eq:pi_optimal}, for a given state $\bm{s}$, we can generate the action $\bm{a}' \sim \pi^*(\cdot|\bm{s})$ using importance sampling. Specifically, $K$ actions $\{\bm{a}^{(k)}_0\}_{k=1}^K$ are sampled using the diffusion-based BC model $\mu_{\bm{\theta}}(\cdot|\bm{s})$ (Line \ref{algs:samplingraw}, Line \ref{algs:samplingbahavior}). The final action $\bm{a}' \sim \pi^*(\cdot|\bm{s})$ is then obtained by applying the importance sampling formula (Line \ref{algs:getfinalaction}):
\begin{equation}
    \begin{aligned}
    \bm{a}' = \sum_{k=1}^K \frac{\mathrm{exp}\left(\alpha Q_{\bm{\phi}}(\bm{s}, \bm{a}^{(k)}_0) \right)}
    {\sum_{k=1}^K \mathrm{exp}\left(\alpha Q_{\bm{\phi}}(\bm{s}, \bm{a}^{(k)}_0) \right)} \bm{a}^{(k)}_0.
    \end{aligned}
    \label{eq:importance}
\end{equation}
This formula computes a weighted combination of actions $\bm{a}^{(k)}_0$ based on their Q-values $Q_\phi(\bm{s}, \bm{a}^{(k)}_0)$, where $\alpha$ is the temperature coefficient that governs the trade-off between adhering to the behavior policy and selecting actions that are more likely to yield higher rewards according to the Q-value function.

Empirically, we observe that selecting the action with the maximum Q-value across the sampled actions (which corresponds to $\alpha=+\infty$) provides more stable performance in many scenarios (Line \ref{algs:getfinalaction}). This is formalized as:
\begin{equation}
    \begin{aligned}
    \bm{a}' = \mathop{\mathrm{argmax}}_{\bm{a}^{(k)}_0} Q_{\bm{\phi}}(\bm{s}, \bm{a}^{(k)}_0) 
    \end{aligned}
    \label{eq:argmax}
\end{equation}
which involves simply choosing the action with the highest Q-value among the \( K \) sampled actions. This approach simplifies the selection process while still leveraging the value function to select high-reward actions.

In both scenarios, our final action is essentially a weighted combination of the actions output by the BC model, which leverages the expressivity and generalization capability of the diffusion model while significantly reducing extrapolation errors introduced by out-of-distribution actions generated. 

Note that  a larger value of \( K \) offers distinct advantages. In the importance sampling case, increasing \( K \) improves the approximation of the target distribution \( \pi^*(\cdot|\bm{s}) \). In the case of selecting the action with the maximum Q-value, a larger \( K \) provides a broader set of options, thus increasing the likelihood of selecting the optimal action. Therefore, we use the DPM-solver \cite{lu2022dpm} as the sampler to enhance sampling efficiency, enabling support for a larger value of \( K \).

\begin{algorithm}[ht]
   \caption{{Action Generation via Diffusion Policy with Critic Guidance}}
   \label{alg:socdgene}
\begin{algorithmic}[1]

   \STATE {\bfseries Input:} State $\bm{s}$, BC model $\mu_{\bm{\theta}}$, Q-networks $Q_{\bm{\phi}}^1,Q_{\bm{\phi}}^2$ {// Input Trained Model and Current State}  \label{algs:inputmodel}
   \FOR{Behavior sampling step $k=1$ {\bfseries to} $K$}
   \STATE Sample $\bm{a}^{(k)}_T$ from $\mathcal{N}(0,\bm{I})$. {// Sampling Raw Noise}\label{algs:samplingraw}
   \STATE Sample $\bm{a}^{(k)}_0$ from BC model $\mu_{\bm{\theta}}(\cdot|\bm{s})$ by solving Equation~\eqref{eq:solve_ode}. {// Diffusion Generation}\label{algs:samplingbahavior}
   \STATE Calculate the Q-value for action $\bm{a}^{(k)}_0$:  $Q_{\bm{\phi}}(\bm{s},\bm{a}^{(k)}_0) = \min_{k=1,2} Q_{\bm{\phi}}^k(\bm{s},\bm{a}^{(k)}_0)$. {// Critic Guidance}  \label{algs:calcq}
   \ENDFOR
   \STATE Obtain final action $\bm{a}'$ via Equation~\eqref{eq:importance} (Importance Sampling) or \eqref{eq:argmax} (Argmax Selecting). {// Obtain Final Action} \label{algs:getfinalaction}
   \STATE {\bfseries Output:} Final action $\bm{a}'$
\end{algorithmic}
\end{algorithm}

\subsubsection{User-level Decomposition} \label{decomposition}
In the MDP of multi-user scheduling systems, the dimensionality of the input state
and action spaces 
scales with the number of users, which poses a considerable challenge to learning in large-scale scenarios due to the restricted capacity of the neural networks used for learning policies.

To address this limitation, we adopt the user-level decomposition techniques introduced in \cite{hu2024multi}. Specifically, we decompose the original MDP into $N$ user-specific sub-MDPs. For each user $i$, the state is $\bm{s}_t^{(i)}=[i, A_i(t), \bm{Q}_i(t), c_i(t)]$, the action is $\bm{a}_t^{(i)}=\bm{v}_i(t)$, and the reward becomes $\bm{r}_t^{(i)}=\omega_i u_i(t) - \bm{v}_i^{T}(t)\bm{Q}_i(t)$ at each time slot $t$. 

By introducing an additional user index \( i \) into the state representation of each sub-MDP, we can train a single model that generates actions for all users. This approach reduces the complexity of the problem by keeping the dimensionality of the input states and actions constant, regardless of the number of users. Consequently, this decomposition enhances the scalability of the algorithm, allowing it to efficiently handle larger multi-user systems.

\subsection{Lagrange Optimization via Offline Data}\label{subsec:lg}
To ensure the resource constraint, SOCD adopts a Lagrangian dual method. According to Remark~\ref{rm:gd}, for a given $\lambda$ value, the gradient of the dual function with respect to  $\lambda$ is: 
\begin{equation}
    \begin{aligned}
        g'(\lambda)=\frac{\partial \mathcal{L}(\pi^{*}(\lambda),\lambda)}{\partial \lambda} = E_0-E_{\pi^{*}(\lambda)}. 
    \end{aligned}
\end{equation}
Here $E_{\pi^{*}(\lambda)}$ is the groundtruth average resource consumption under the policy $\pi^{*}(\lambda)$. 
Thus, we can update the Lagrange multiplier $\lambda$ according to: 
\begin{equation}
\label{eq:aorarlambda}
    \begin{aligned}
        \lambda \leftarrow \lambda - \alpha g'(\lambda) = \lambda - \alpha (E_0-E_{\pi^{*}(\lambda)}).
    \end{aligned}
\end{equation}
Then, after updating the value of $\lambda$, we modify the offline dataset's reward using $r_t=D(t)-\lambda E(t)$ and iterate the learning procedure.

While our method is similar to that in \cite{hu2024multi}, the offline learning nature means that we cannot learn the value \( E_{\pi^{*}(\lambda)} \) by interacting with the environment, as is done in online algorithms \cite{hu2024multi}.

To address the problem of estimating $E_{\pi^{*}(\lambda)}$, we propose to estimate the resource consumption in an offline manner. Specifically, given a fixed $\lambda$, we train the algorithm and obtain a policy. Then, we sample $n$ trajectories $\{\mathcal{T}_j\}_{j=1}^{n}$ from the dataset $\mathcal{D}$. For each trajectory $\mathcal{T}_j=(\bm{s}^j_1,\bm{a}^j_1,...\bm{s}^j_t,\bm{a}^j_t,...,\bm{s}^j_T,\bm{a}^j_T)$, we feed the states $\bm{s}^j_t$ into the network, and obtain the output action $\hat{\bm{a}}^j_t = [\hat{\bm{v}}^j_1(t), ..., \hat{\bm{v}}^j_N(t)]\sim\pi_{\bm{\theta}}(\cdot|\bm{s}^j_t)$. According to~\eqref{eq:resource}, the resource consumption is estimated as $\hat{E}^j_t=\sum_{i=1}^N \hat{\bm{v}}^j_i(t)^{\top} \bm{Q}^j_i(t)$. Here $\bm{Q}^j_i(t)$ is the buffer information included in $\bm{s}_t^j$. Thus, the average resource consumption over the sampled trajectories can be given as 
\begin{equation}
    \begin{aligned}
        \hat{E}_{\pi^{*}(\lambda)}=\frac{1}{n}\sum_{j=1}^{n}\frac{1}{T}\sum_{t=1}^{T}\hat{E}^j_t.
    \end{aligned}
    \label{eq:estimate}
\end{equation}

We use Equation~\eqref{eq:estimate} as an estimate for $E_{\pi^{*}(\lambda)}$. This approach effectively avoids interacting with the environment and enables an offline estimation of $E_{\pi^{*}(\lambda)}$. {Note that if the policy is allowed to interact with the environment, we can evaluate the policy of SOCD in the online phase to obtain the groundtruth value of $E_{\pi^{*}(\lambda)}$ and update the Lagrange multiplier more accurately. However, in the offline phase, it is impossible to acquire the accurate value.} 

{The pseudocode for the SOCD algorithm is shown in Algorithm \ref{alg:socd}. 
Our algorithm provides a systematic way to integrate the offline RL algorithm to solve the offline network scheduling problems. 
The novelty of the algorithm lies in two key aspects: First, SOCD integrates a diffusion-based scheduling policy into the framework, leveraging its superior performance in general offline RL tasks \cite{chen2023score}. Second, SOCD requires training the behavioral cloning (BC) model, specifically the diffusion policy, only once in the offline scheduling task. This design significantly enhances training efficiency compared to baseline algorithms such as SOLAR \cite{li2024offline}, which necessitate multiple policy training iterations following each update of the Lagrange multiplier. In contrast, SOCD completely eliminates the need for iterative policy updates, requiring the diffusion-based policy to be trained only once.}

\begin{algorithm}[ht]
   \caption{\underline{S}cheduling By \underline{O}ffline Learning with \underline{C}ritic Guidance and \underline{D}iffusion Generation ($\mathtt{SOCD}$)}
   \label{alg:socd}
\begin{algorithmic}[1]
   
   \STATE {\bfseries Input:} Dataset $\mathcal{D}$, initialize Diffusion-based BC model $\mu_{\bm{\theta}}$ with random parameter $\bm{\theta}$ {// BC Model Initialization}  \label{algs:bcinitialization}
   \FOR{BC training step $b=1$ {\bfseries to} $S$}
   \STATE Sample a random minibatch of $\mathcal{B} \times T$ state-action Pairs $(\bm{s},\bm{a})$ from dataset $\mathcal{D}$. {// Sampling State-action pairs}\label{algs:samplingsapair}
   \STATE Update BC Model $\mu_{\bm{\theta}}$ to minimize \eqref{eq:train_dif}. {// Update BC Model} \label{algs:updatebc}
   \ENDFOR
   \FOR{Lagrange multiplier iteration step $k=1,...,K$}
   \STATE {\bfseries Input:} Initialize Q-networks $Q_{\bm{\phi}}^1,Q_{\bm{\phi}}^2$, policy network $\pi$ with random parameters $\bm{\phi}^1,\bm{\phi}^2$, target Q-networks with $\overline{\bm{\phi}}^1 \leftarrow \bm{\phi}^1,\overline{\bm{\phi}}^2 \leftarrow \bm{\phi}^2$,  Lagrange multiplier $\lambda$. {// Critic Initialization} \label{algs:criticinitialization}
   \FOR{training step $i=1$ {\bfseries to} $T$}
   \STATE Sample a random minibatch of $\mathcal{B}$ trajectories $\left\{\mathcal{T}_j=(\bm{s}^j_1,\bm{a}^j_1,\bm{r}^j_1,...\bm{s}^j_t,\bm{a}^j_t,\bm{r}^j_t,...,\bm{s}^j_T,\bm{a}^j_T,\bm{r}^j_T)\right\}_{j=1}^{\mathcal{B}}$ from dataset $\mathcal{D}$ with reward updated (${r}=E-\lambda D$). {// Sampling Trajectories}\label{algs:samplingtraj}
   \STATE Update critics $\bm{\phi}^1,\bm{\phi}^2$ to minimize \eqref{eq:criticloss}. {// Update Critic} \label{algs:updatecritics}
   \STATE Update target networks: $\overline{\bm{\phi}}^k\leftarrow\rho\bm{\phi}^k+(1-\rho)\overline{\bm{\phi}}^k$,$(k=1,2)$. {// Soft Update Target Critic}\label{algs:softupdate}
   \ENDFOR
   \STATE Sample sequential actions via Algorithm~\ref{alg:socdgene}, obtain the value $\hat{E}_{\pi^{*}(\lambda)}$ via \eqref{eq:estimate} and update $\lambda$ via \eqref{eq:aorarlambda}. {// Update Lagrange Multiplier}\label{algs:updatelambda}
   \ENDFOR
\end{algorithmic}
\end{algorithm}

\section{Experimental Results}\label{sec:exp}

We evaluate SOCD in numerically simulated environments. Section~\ref{subsec:es} details the experimental setup and the properties of the offline dataset. Section~\ref{exp:nw} describes the network architecture and provides the hyperparameter configurations. In Sections~\ref{subsec:baselines} and~\ref{subsec:er}, we introduce several baseline scheduling algorithms known for their effectiveness, and compare SOCD's performance against them in both standard and more demanding offline scheduling scenarios.

\subsection{Environment Setup}\label{subsec:es}
This subsection specifies the setup of the delay-constrained network environments, as depicted in Figure~\ref{fig:sm}, where our experiments are conducted. We employ a setup similar to existing works \cite{singh2019throughput, chen2018timely, hu2024multi}, where each interaction step corresponds to a single time slot in the simulated environment. A total of $T=100$ interaction steps (or time steps) define one episode during data collection and training. Each episode is represented by the data sequence $(\bm{s}_0, \bm{a}_0, \dots, \bm{s}_T, \bm{a}_T)$. In addition, we perform online algorithm evaluations over 20 rounds, each consisting of 1000 time units of interaction with the environment, to ensure accuracy and reliability.

To showcase SOCD's superior performance across varying system dynamics, we conduct experiments in a variety of diverse environments. The specific settings for all environments used in our experiments are detailed in Table~\ref{tab:env} and Table~\ref{tab:envdetail}.\footnote{Here, $[a \pm b]^n$ denotes an $n$-dimensional vector with mean $a$ and standard deviation $b$.} 

\begin{table}[htbp]
\centering
	\caption{Environment Information: The table outlines various environment configurations used in this study. ``Poisson" refers to a Poisson arrival process and Markovian channel conditions, while ``Real" refers to environments based on real records. ``-$x$hop" indicates the number of hops in the network, while ``-$x$user" denotes the number of users in the environment (without this flag the default is 4 users). ``-partial" signifies the presence of partial information in the environment.}

	\label{tab:env}
 \resizebox{0.48\textwidth}{!}{
	\begin{tabular}{cccccc}
		\toprule
		Environment & User & Flow Generation & Channel Condition & Hop & Partial \\
		\midrule
            Poisson-1hop & 4 & Poisson & Markovian & 1 & False \\
            Poisson-2hop & 4 & Poisson & Markovian & 2 & False \\
            Poisson-3hop & 4 & Poisson & Markovian & 3 & False \\
            Poisson-100user & 100 & Poisson & Markovian & 1 & False \\
            Real-1hop & 4 & Real Record & Real Record & 1 & False \\
            Real-2hop & 4 & Real Record & Real Record & 2 & False \\
            Real-partial & 4 & Real Record & Real Record & 1 & True \\
		\bottomrule
	\end{tabular}}
\end{table}

\begin{table}[htbp]
\centering
	\caption{Environment Information Continued. The environment characteristics follow the same explanation as in Table~\ref{tab:env}.}
	\label{tab:envdetail}
 \resizebox{0.48\textwidth}{!}{
	\begin{tabular}{cccc}
		\toprule
		Environment & Delay $\tau$ & Weight $\omega_i$ & Instantaneous Arrival Rate \\
		\midrule
            Poisson-1hop & [4, 4, 4, 6] & [4, 2, 1, 4] & [3, 2, 4, 2] \\
            Poisson-2hop & [5, 5, 5, 7] & [4, 2, 1, 4] & [3, 2, 4, 2] \\
            Poisson-3hop & [6, 6, 6, 8] & [4, 2, 1, 4] & [3, 2, 4, 2] \\
            Poisson-100user & $[3.39\pm 1.75]^{100}$ & $[3.09\pm1.46]^{100}$ & $[3.64\pm1.18]^{100}$ \\
            Real-1hop &  [2, 5, 1, 3] & [1, 4, 4, 4] & / \\
            Real-2hop & [3, 6, 2, 4] & [1, 4, 4, 4] & / \\
            Real-partial &  [2, 5, 1, 3] & [1, 4, 4, 4] & / \\
		\bottomrule
	\end{tabular}}
\end{table}
\subsubsection{Arrivals and Channel Conditions}
The arrivals and channel conditions are modeled in two distinct ways: (\romannum{1}) The arrivals follow the Poisson processes, while the channel conditions are assumed to be Markovian. (\romannum{2}) The arrivals are based on trajectories with various types from an LTE dataset \cite{loi2018predict}, which records the traffic flow of mobile carriers' 4G LTE network over the course of approximately one year. The channel conditions are derived from a wireless 2.4GHz dataset \cite{taotao2021wireless}, which captures the received signal strength indicator (RSSI) in the check-in hall of an airport.

\subsubsection{Outcomes}\label{oc}
We adopt the same assumption for the probability of successful service as proposed in \cite{chen2018timely, hu2024multi}. Specifically, the probability of successful service for user $i$, given channel condition $c$ and allocated resource $v$ 
is modeled as follows:
\begin{equation}\label{eq:so}
    P_i(v,c) = \frac{2}{1 + \exp\left( -\frac{2v}{l_i^{3}c} \right)} - 1,
\end{equation}
where $l_i$ denotes the distance between user $i$ and the server. This model is representative of a wireless downlink system, where $v$ is interpreted as the transmission power of the antenna responsible for transmitting the current packet \cite{chen2018timely}.

\subsubsection{Offline Dataset}
We set the episode length to $T=100$. Each dataset comprises $J=5000$ episodic trajectories, generated using the online RSD4 algorithm \cite{hu2024multi}. During the training procedure of RSD4, the algorithm interacts with the environment to generate new trajectories, which are then stored in the replay buffer for training purposes. We retain the first $5000$ trajectories from the RSD4 replay buffer to form the offline dataset. Details of this dataset are provided in Table~\ref{tab:data}, all of which are of medium quality, i.e., generated by a suboptimal behavior policy.

\begin{table}[htbp]
\centering
	\caption{Dataset Information. The environment characteristics follow the same explanation as in Table~\ref{tab:env}. L.R. refers to the Learning Rate. }
	\label{tab:data}
 \resizebox{0.48\textwidth}{!}{
	\begin{tabular}{cccccc}
		\toprule
		Environment & RSD4 L.R. & $\lambda$ under RSD4 & Throughput & Resource Consumption \\
		\midrule
            Poisson-1hop & 0.0001 & 1.0 & 12.77$\ \pm \ $1.14 & 14.64$\ \pm \ $4.47 \\
            Poisson-2hop & 0.0001 & 0.5 & 9.52$\ \pm \ $1.44 & 15.71$\ \pm \ $4.61  \\
            Poisson-3hop & 0.0001 & 0.001 & 1.62$\ \pm \ $0.44 & 25.90$\ \pm \ $2.27  \\
            Poisson-100user & 0.0003 & 0.1 & 53.55$\ \pm \ $4.12 & 356.46$\ \pm \ $102.83  \\
            Real-1hop & 0.0001 & 1.0 & 5.84$\ \pm \ $1.56 & 6.70$\ \pm \ $2.54  \\
            Real-2hop & 0.0001 & 0.1 & 0.56$\ \pm \ $0.47 & 5.26$\ \pm \ $4.13  \\
            Real-partial & 0.0001 & 1.0 & 5.94$\ \pm \ $1.23 & 5.60$\ \pm \ $1.63  \\
		\bottomrule
	\end{tabular}}
\end{table}

\subsection{Network Structures and Hyperparameters}\label{exp:nw}
For the score model $\mathbf{s}_{\bm{\theta}}(\cdot,\bm{s},t)$ used in the diffusion-based behavior cloning (BC) component of our SOCD algorithm as defined in Equation~\eqref{eqreq:invsde}
, we embed the time slot $t$ via Gaussian Fourier Projection \cite{rahimi2007random}, resulting in a latent representation of dimension 32, while the state $\bm{s}$ is embedded as a latent variable of dimension 224. These embeddings are concatenated and subsequently passed through a 4-layer multi-layer perceptron (MLP) with hidden layer sizes of $[512, 256, 256, 256]$, using the SiLU activation function \cite{ramachandran2017swish}. For the critic model, we employ a 2-layer MLP with 256 hidden units and ReLU activation. The hyperparameters of the SOCD algorithm are detailed in Table~\ref{tab:hyper}.

\begin{table}[htbp]
\centering
	\caption{Hyperparameter Information.}
	\label{tab:hyper}
 \resizebox{0.48\textwidth}{!}{
	\begin{tabular}{cl}
		\toprule
		Hyperparameter & Value  \\
		\midrule
            BC Model Training Steps & $1 \times 10^5$ \\
            BC Model Learning Rate & $1\times 10^{-4}$\\
            BC Model Optimizer & Adam \\
            Critic Training Steps & $3 \times 10^3$ \\
            Critic Learning Rate & $3\times 10^{-4}$\\
            Critic Optimizer & Adam \\
            Soft Update Parameter $\rho$ & $0.05$\\
            Discount Factor $\gamma$ & $0.8$\\
            Temperature Coefficient $\alpha$ & $100$\\
            DPM-solver Sampling Steps & $10$\\
            Number of Behavior Actions sampled per state & $1024$\\
            $\omega_{min}$ & 0.1 \\
            $\omega_{max}$ & 20 \\
            Batch Size $\mathcal{B}$ & 50 \\
		\bottomrule
	\end{tabular}}
\end{table}
The network architecture and hyperparameters for the SOLAR algorithm used in this study are consistent with those presented in \cite{li2024offline}.

\subsection{Baselines} \label{subsec:baselines}
We compare our SOCD algorithm with several alternative algorithms, including the offline DRL algorithm Behavior Cloning (BC) \cite{lange2012batch}, a DRL-based scheduling algorithm SOLAR \cite{li2024offline}, and traditional scheduling methods including Uniform and Earliest Deadline First (EDF) \cite{elsayed2006channel}, as described below. Note that while there exist other online DRL algorithms, e.g., RSD4 \cite{hu2024multi}, they require online interaction and are thus not suitable for the offline learning setting. We also exclude the programming-type methods from our baselines, as they requires access to full environment information, whereas in the offline setting, dynamic information is completely unavailable. 

\paragraph{Behavior Policy}
We compare the performance with that of the behavior policy, which is represented by the average score of all trajectories in the offline dataset.  

\paragraph{Uniform}
This method allocates available resources uniformly across all packets in the buffer, without considering their individual characteristics or deadlines.

\paragraph{Earliest Deadline First (EDF)}
The EDF algorithm prioritizes packets based on the remaining time to their respective deadlines \cite{elsayed2006channel}. Specifically, it serves packets with the shortest remaining time to deadline first, assigning the maximum available resource allocation, $e_{max}$, to these packets. The process is repeated for packets with the next smallest remaining time to deadline, continuing until the total resource allocation reaches the hard limit $E_{0}$.

\paragraph{Behavior Cloning (BC)}
The Behavior Cloning method aims to learn a policy network by minimizing the following objective \cite{lange2012batch}:
\begin{equation}
\mathcal{L}(\theta) = \mathbb{E}_{(\bm{s},\bm{a}) \sim \mathcal{D}, \tilde{\bm{a}} \sim \pi_{\theta}(\cdot|\bm{s})} \left[ (\tilde{\bm{a}} - \bm{a})^2 \right].
\end{equation}
This objective strives to imitate the behavior policy by reducing the discrepancy between the actions predicted by the policy network and the actions observed in the dataset. In our experiments, we leverage the diffusion-based BC model within our SOCD algorithm to evaluate the performance of the Behavior Cloning approach. 

\paragraph{SOLAR} 
{SOLAR is a scheduling algorithm based on offline reinforcement learning \cite{li2024offline}, where the RL maximizer is an actor-critic RL framework with a Gaussian policy. The critic is trained using a conservative regularized loss function, while the actor loss incorporates Actor Rectification \cite{pan2022plan} technique, as detailed in \cite{li2024offline}. Although both SOLAR and our SOCD algorithm share the same framework, comprising an offline RL maximizer and Lagrange multiplier optimization, SOLAR exhibits inferior performance compared to SOCD. This performance is primarily due to the higher fidelity and the expressivity provided by the diffusion model within SOCD, as demonstrated in the results below.}

\subsection{Performance Evaluation} \label{subsec:er}

{To demonstrate the effectiveness of SOCD, we evaluate its performance across a diverse set of environments, as summarized in Table \ref{tab:env}. These environments are carefully designed to capture a broad range of challenges, including single-hop and multi-hop network configurations, both simulated and real-world arrival and channel conditions, settings with full and partial observability, as well as large-scale environments that test the scalability and robustness of the proposed method. }

{In our evaluation, we focus on Weighted Throughput and Resource Consumption as the primary performance indicators. It is important to note that under our strict delay constraint model \cite{li2024offline}, any packet not served within its deadline is immediately discarded. Consequently, the Throughput metric directly quantifies the Packet Delivery Ratio (PDR) and delay satisfaction, as it exclusively accounts for valid, non-expired jobs. Furthermore, we evaluate constraint adherence by explicitly comparing the Actual Resource Consumption against the target Resource Constraint, providing a direct visual measure of the constraint-violation rate. In the following sections, we present and discuss the results in detail.}

\subsubsection{Single-hop Environments}
We begin by evaluating the performance in a single-hop environment, where both arrivals and channel conditions are modeled as Poisson processes, as denoted by {Poisson-1hop} in Table \ref{tab:env}. 
\begin{figure}[htbp]
	\centering
\includegraphics[width=\columnwidth]{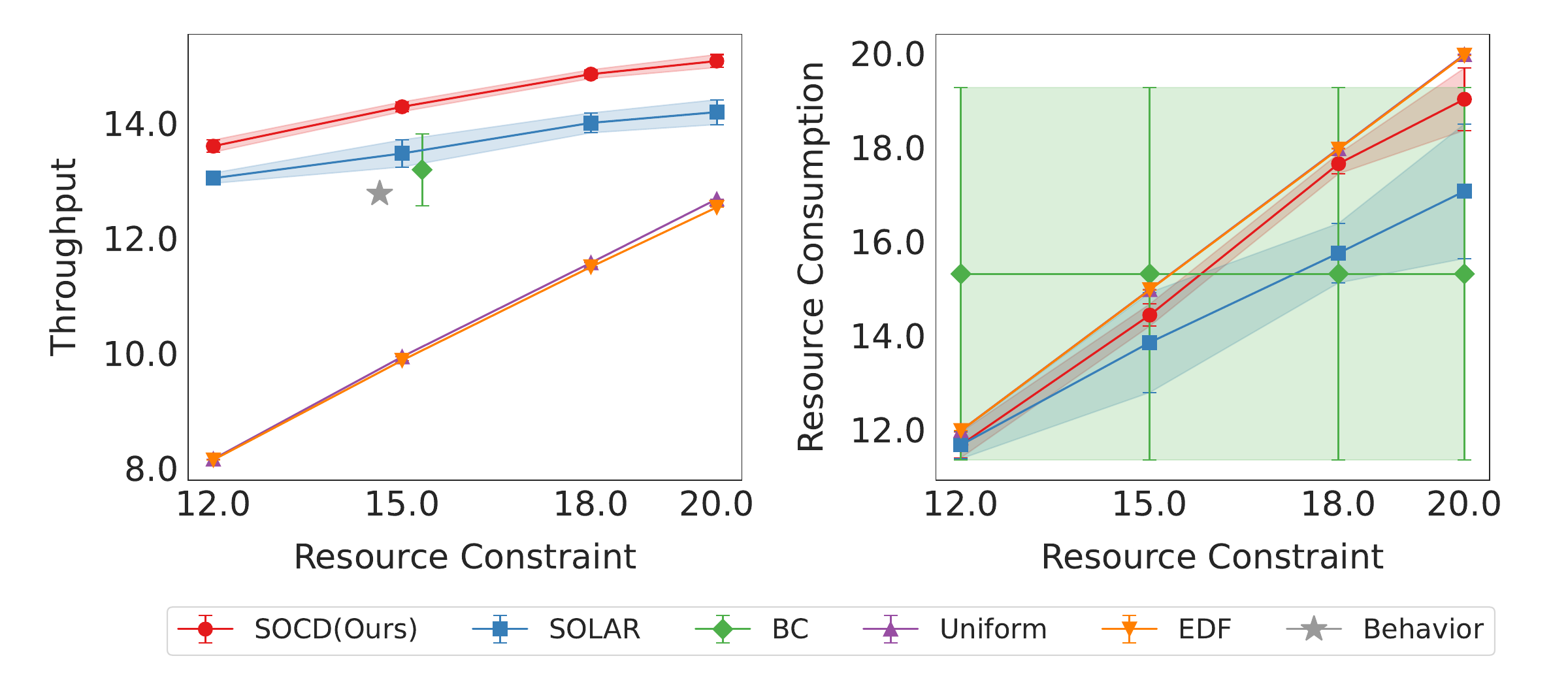}
	\caption{Comparison of different algorithms in the {Poisson-1hop} environment, where arrivals and channel conditions are modeled as Poisson processes. The plots display throughput (left) and resource consumption (right) under varying resource constraints.}
	\label{fig:pois}
\end{figure}

The results are shown in Figure~\ref{fig:pois}. The Behavior Cloning (BC) policy consistently exhibits the same variance because its training relies solely on the state and action data from the dataset, without incorporating the reward signal which includes the Lagrange multiplier. As a result, its outcomes remain unaffected by resource constraints.

The results demonstrate that SOCD effectively learns an efficient policy, outperforming traditional scheduling methods, such as Uniform and EDF.
Additionally, both SOCD and SOLAR outperform the behavior policy, indicating that DRL-based algorithms can successfully generalize beyond the suboptimal behavior policy embedded in the offline dataset to effectively utilize the offline data to learn improved policies. Notably, SOCD consistently outperforms SOLAR in terms of both throughput and resource consumption efficiency, further highlighting the superiority of SOCD.

\subsubsection{Multi-hop Environments}
We extend the evaluation to multi-hop settings, starting with {Poisson-2hop}, a 2-hop network environment where arrivals and channel conditions are modeled as Poisson processes.

\begin{figure}[htbp]
	\centering
        \includegraphics[width=\columnwidth]{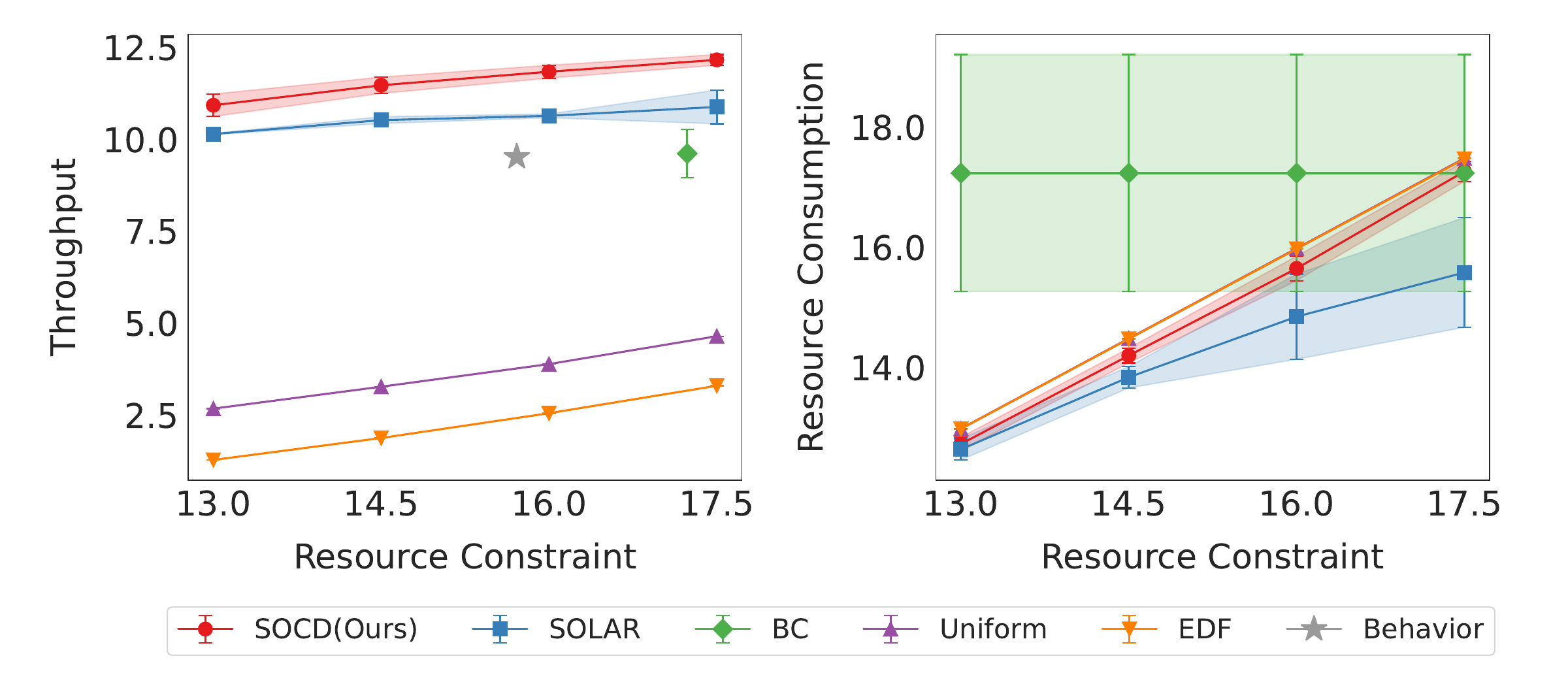}
	\caption{Comparison of different algorithms in a 2-hop environment, {Poisson-2hop}, with arrivals and channel conditions modeled as Poisson processes. The left column shows throughput, while the right column illustrates resource consumption under varying resource constraints.}
	\label{fig:pois_2hop}
\end{figure}

As shown in Figure~\ref{fig:pois_2hop}, SOCD outperforms SOLAR, further demonstrating its robustness in optimizing scheduling policies across different network topologies. A noteworthy point is that the BC policy and the behavior policy exhibit significantly different performance, suggesting that the learned behavior policy can diverge from the average policy observed in the offline dataset due to the diversity in actual behavior. Despite this, SOCD consistently delivers superior performance compared to BC, highlighting the critical role of Critic Guidance.

\subsubsection{Real Data Simulation}
{Building on the insights gained from single-hop and multi-hop environments, we now evaluate the {transferability and robustness} of our algorithms in more challenging scenarios, where arrivals and channel conditions are drawn from real-world datasets. Unlike synthetic Poisson-based environments, where the scheduling task is relatively straightforward, real-world traces introduce significant noise, irregularities, and distributional shifts. This setup serves as a critical testbed for verifying whether the policy trained on medium-quality simulated data can effectively generalize to the complex dynamics of actual systems. Specifically, {Real-1hop} and {Real-2hop}, as described in Table~\ref{tab:env}, represent environments where the arrival processes are modeled using a 4G LTE dataset \cite{loi2018predict}, while the channel conditions are derived from a 2.4GHz wireless dataset \cite{taotao2021wireless}.}

\begin{figure}[htbp]
    \centering
    
    \begin{subfigure}[c]{\columnwidth}
        \centering
        \includegraphics[width=\columnwidth]{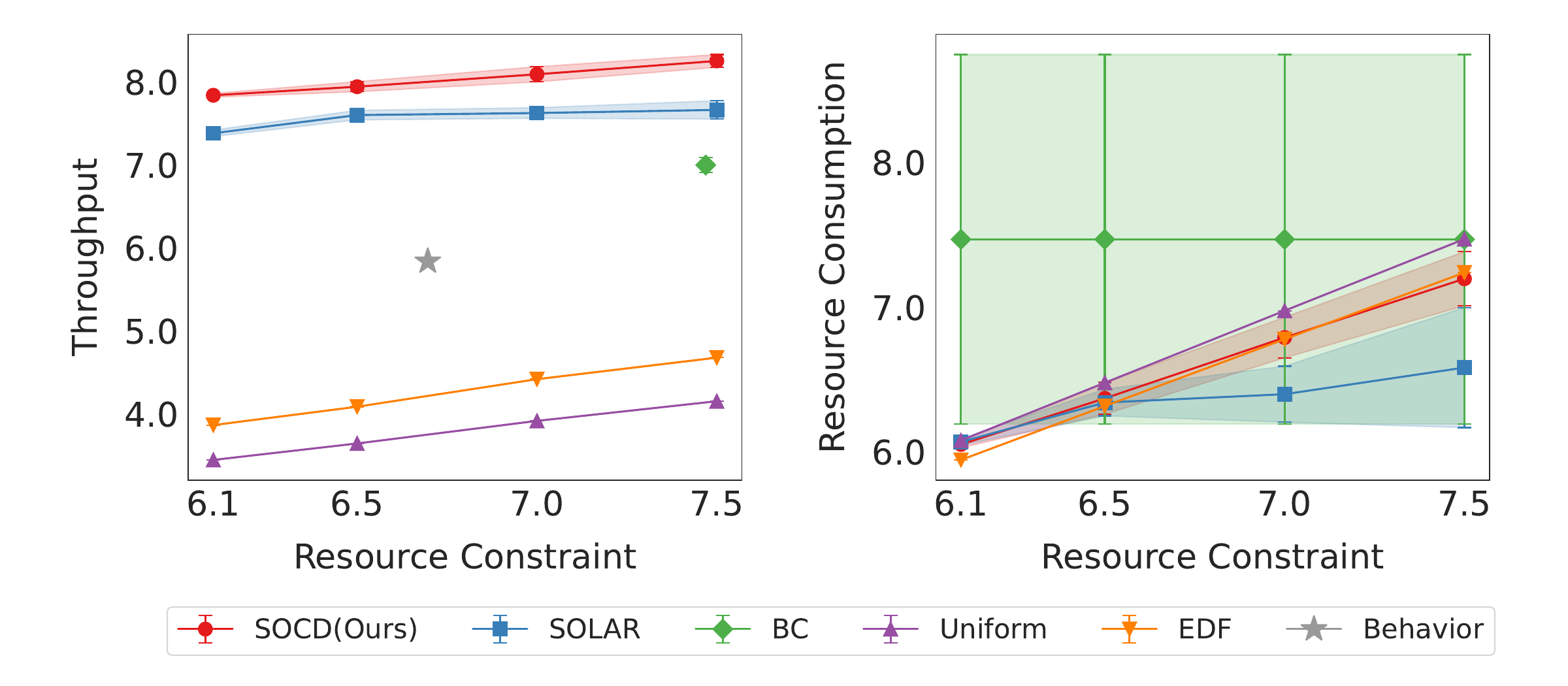}
        \caption{Comparison of different algorithms in {Real-1hop}.}
        \label{fig:real}
    \end{subfigure}
    
    \vspace{0.3cm}
    
    \begin{subfigure}[c]{\columnwidth}
        \centering
        \includegraphics[width=\columnwidth]{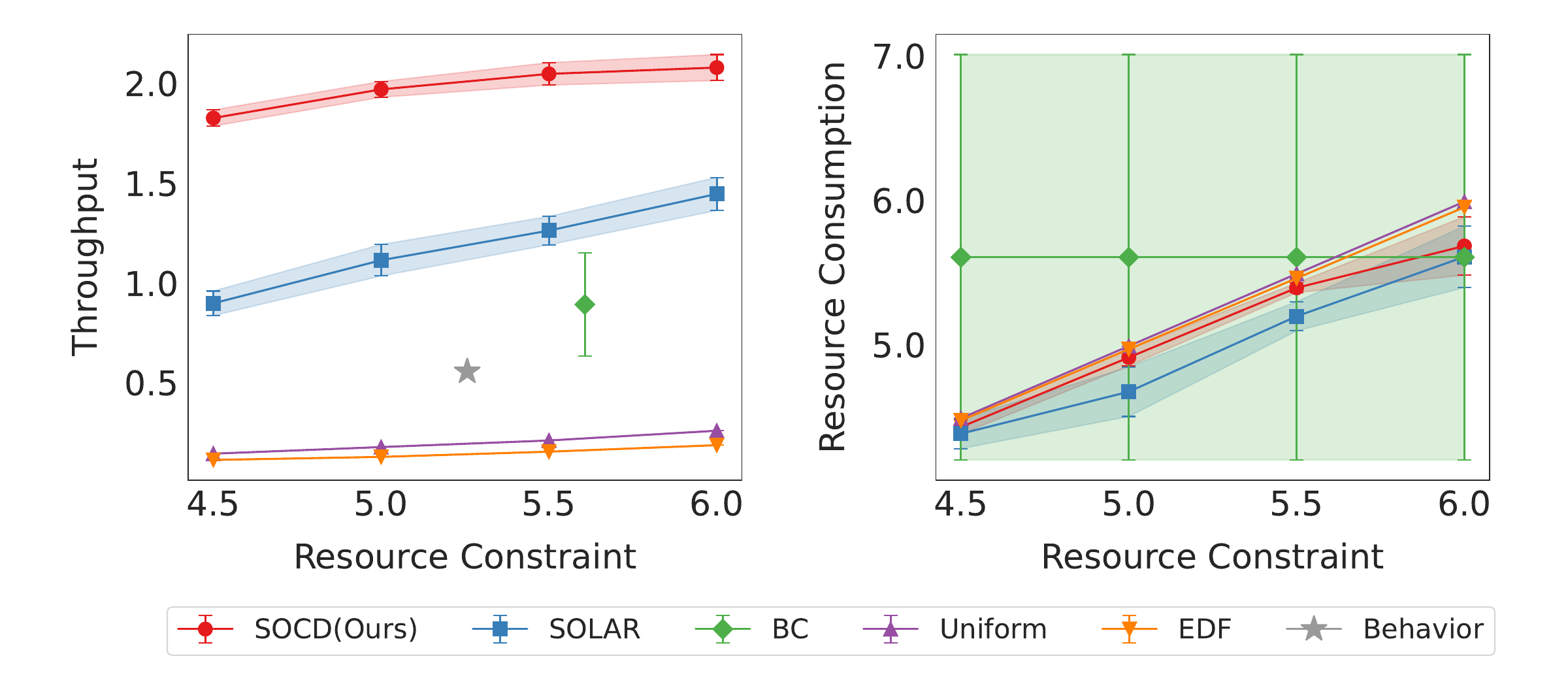}
        \caption{Comparison of different algorithms in {Real-2hop}.}
        \label{fig:real_2hop}
    \end{subfigure}

    \caption{Comparison of different algorithms in two environments that utilize real-world data to model arrivals and channel conditions. {Real-1hop} (upper) is a single-hop environment, while {Real-2hop} (lower) is a 2-hop environment. The left column shows throughput, and the right column illustrates resource consumption under varying resource constraints.}
    \label{fig:reals}
\end{figure}

The results presented in Figure~\ref{fig:reals} underscore the consistent superiority of SOCD over other algorithms, even in these more complex, real-data-driven environments. This highlights SOCD’s robustness and its ability to optimize scheduling policies despite the unpredictability and complexity inherent in real-world network data. Notably, under resource constraints of 6.1 and 6.5 in {Real-1hop} (Figure~\ref{fig:real}) and 4.5 in {Real-2hop} (Figure~\ref{fig:real_2hop}), the resource consumption of SOCD and SOLAR is nearly identical. However, SOCD achieves notably higher throughput than SOLAR, demonstrating its superior resource utilization efficiency in terms of algorithm performance.

\subsubsection{Dynamic Changes}
In this section, we further evaluate the performance of the algorithms in three more challenging environments characterized by different dynamic conditions:
(\romannum{1}) a {Real-partial} scenario, where channel conditions are unavailable during decision-making, adding the challenge of partial observability; 
(\romannum{2}) a {Poisson-3hop} scenario, which models a 3-hop network and introduces increased complexity due to the additional hop; 
and (\romannum{3}) a {Poisson-100user} scenario, which simulates a high user density and requires scalable solutions to manage the increased load and complexity.

\paragraph{Partially Observable Systems}
We begin by evaluating the algorithms in a partially observable environment Real-partial, where the channel condition is unavailable for decision-making. Despite the limited observability, SOCD not only outperforms SOLAR in equivalent conditions but also nearly matches the performance of SOLAR in fully observable settings, as evidenced by the throughput under resource constraints of 6.5 and 7.0 (Figure~\ref{fig:real} and Figure~\ref{fig:partial}). This highlights SOCD’s exceptional generalization capability, enabling it to perform well even when it has no knowledge of the hidden factors driving the dynamic environment.
\begin{figure}[htbp]
	\centering
    \includegraphics[width=\columnwidth]{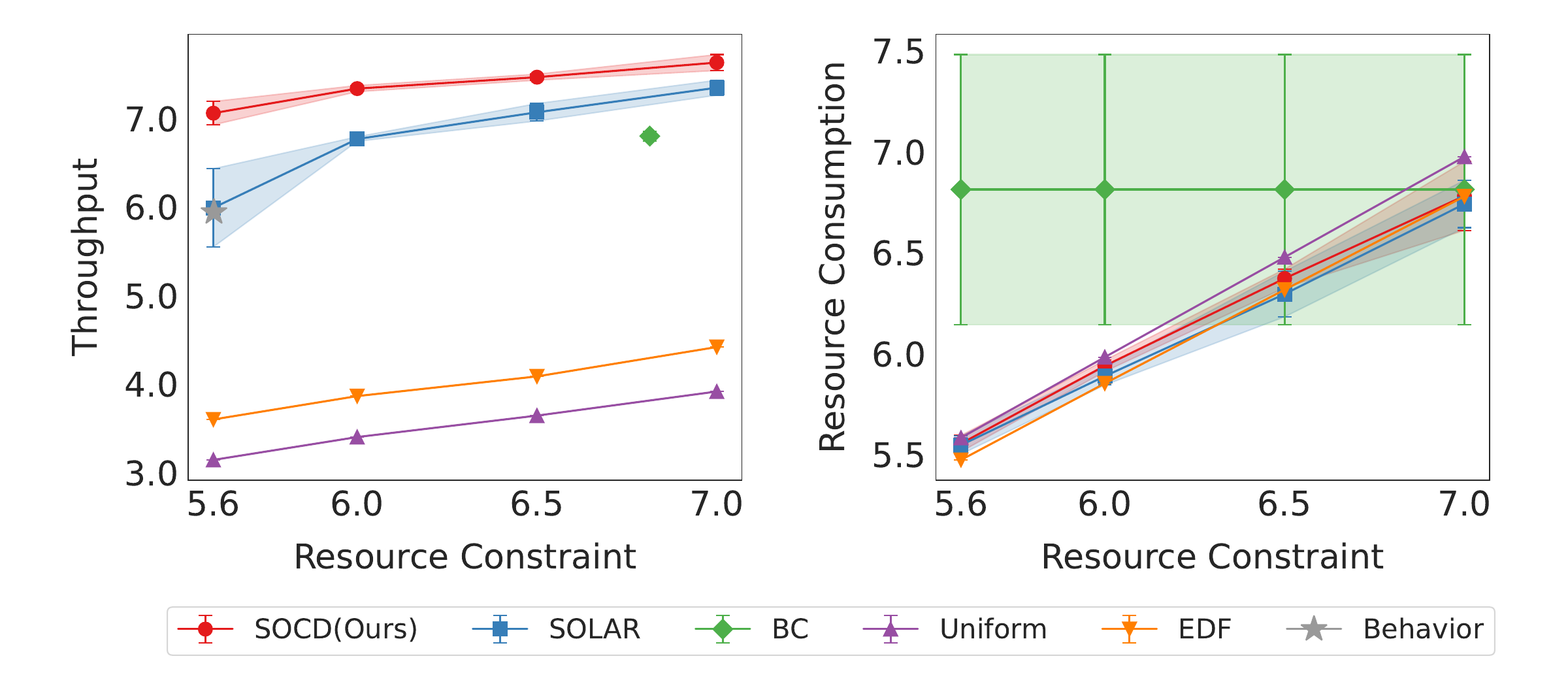}
	\caption{Comparison of different algorithms in a partially observable environment, where the channel condition is unobservable. The left plot shows throughput, and the right plot  illustrates resource consumption under varying resource constraints.}  
	\label{fig:partial}
\end{figure}

\paragraph{More Hops}
As the number of hops increases, the achievable throughput for a given amount of resource decreases, and throughput becomes dependent on a broader range of factors. This complicates the establishment of a clear correlation between resource consumption and throughput, making the scheduling task more challenging. Consequently, it becomes increasingly difficult for the policy to fully utilize the available resource under higher resource constraints. Specifically, SOLAR, which relies on a Gaussian model to approximate the policy, struggles to adapt to higher resource constraints, as evidenced by the right plot in Figure~\ref{fig:3hop}.

In contrast, SOCD employs a diffusion model for behavior cloning, ensuring that its performance is never worse than the behavior policy, which provides a solid foundation. Building upon this, SOCD incorporates Critic Guidance to consistently enhance its performance. As shown in Figure~\ref{fig:3hop}, it effectively utilizes resource to improve throughput, even as the number of hops increases.
\begin{figure}[htbp]
	\centering
    \includegraphics[width=\columnwidth]{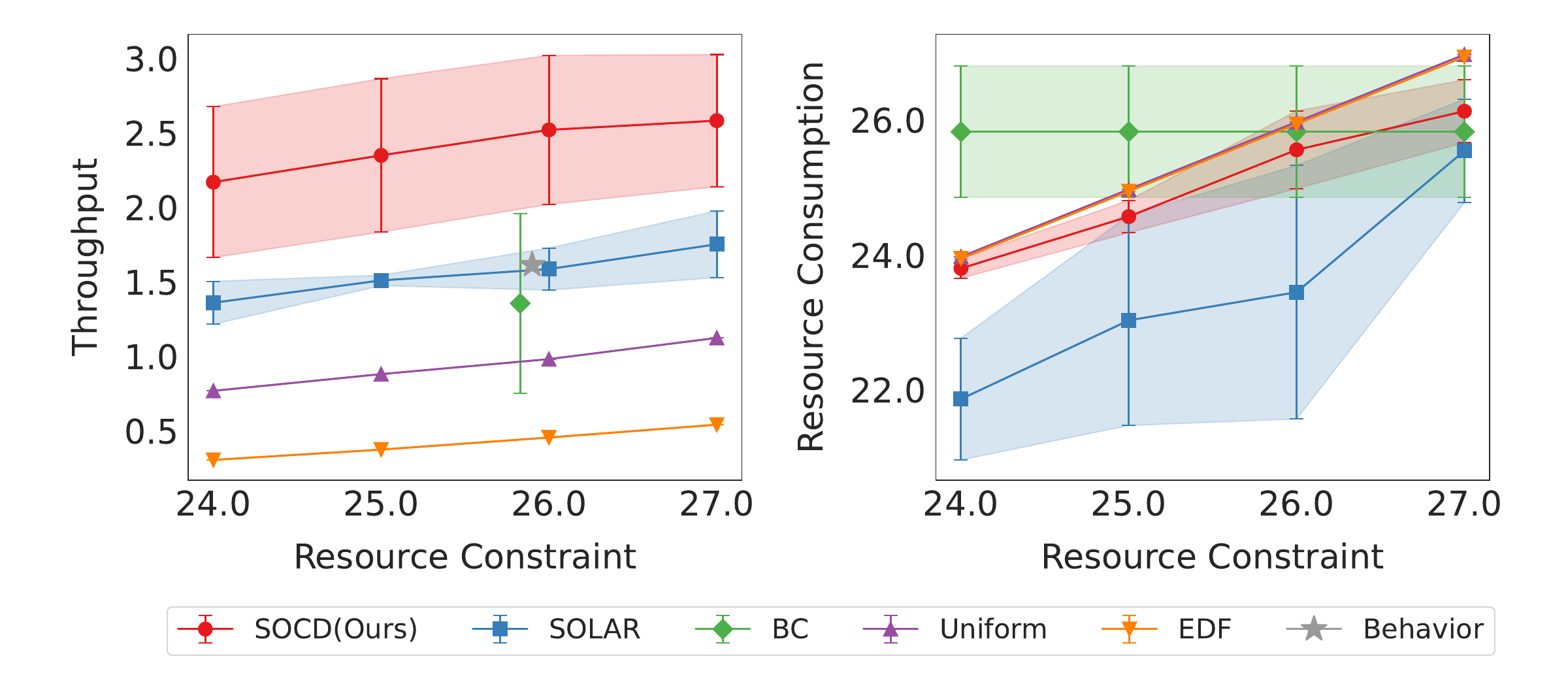}
	\caption{Comparison of different algorithms as the number of hops increases to 3. The left plot shows the throughput, while the right plot illustrates resource consumption under varying resource constraints.}
	\label{fig:3hop}
\end{figure}

\paragraph{Scalability in User Number} 
As the number of users grows to $100$, scalability becomes a significant challenge due to the rapid increase in the dimensionality of the input state and action spaces, which leads to a substantial performance degradation. To address this, SOLAR, BC, and SOCD are evaluated with their user-level decomposition versions as discussed in Section~\ref{decomposition}, which is unnecessary in four-user networks but vital in large-scale scenarios. This is because, with fewer users, the size of the state and action space is small enough for a standard network to handle, but with more users, the required network size expands to an unbearable degree, necessitating decomposition. A higher user density also introduces additional challenges, particularly in terms of stability. As the number of users grows, the variability in scheduling decisions increases, and the system becomes more prone to performance fluctuations. This leads to larger deviations in both throughput and resource consumption, making it difficult to maintain consistent performance over time.

As shown in Figure~\ref{fig:100user}, traditional scheduling methods such as EDF and Uniform effectively utilize the entire resource capacity, but fail to deliver competitive throughput. Despite learning the policy in a single round without the use of Lagrange optimization, BC performs similarly to SOLAR in terms of throughput, highlighting the strong expressivity of the diffusion model. This allows BC to capture subtle variations in user index $i$ within the state space, generating more diverse scheduling strategies for each individual user. However, BC exhibits higher resource consumption and significantly larger variance, with the upper bound of its resource consumption consistently exceeding the constraint.

By incorporating Critic Guidance on top of BC, SOCD addresses these limitations and significantly outperforms both SOLAR and BC. SOCD’s throughput is consistently superior to both BC and SOLAR, and its resource variance remains well within the constraint, even at higher user densities. The superior stability in throughput and more consistent resource consumption underscore SOCD's clear advantage.
\begin{figure}[htbp]
	\centering
    \includegraphics[width=\columnwidth]{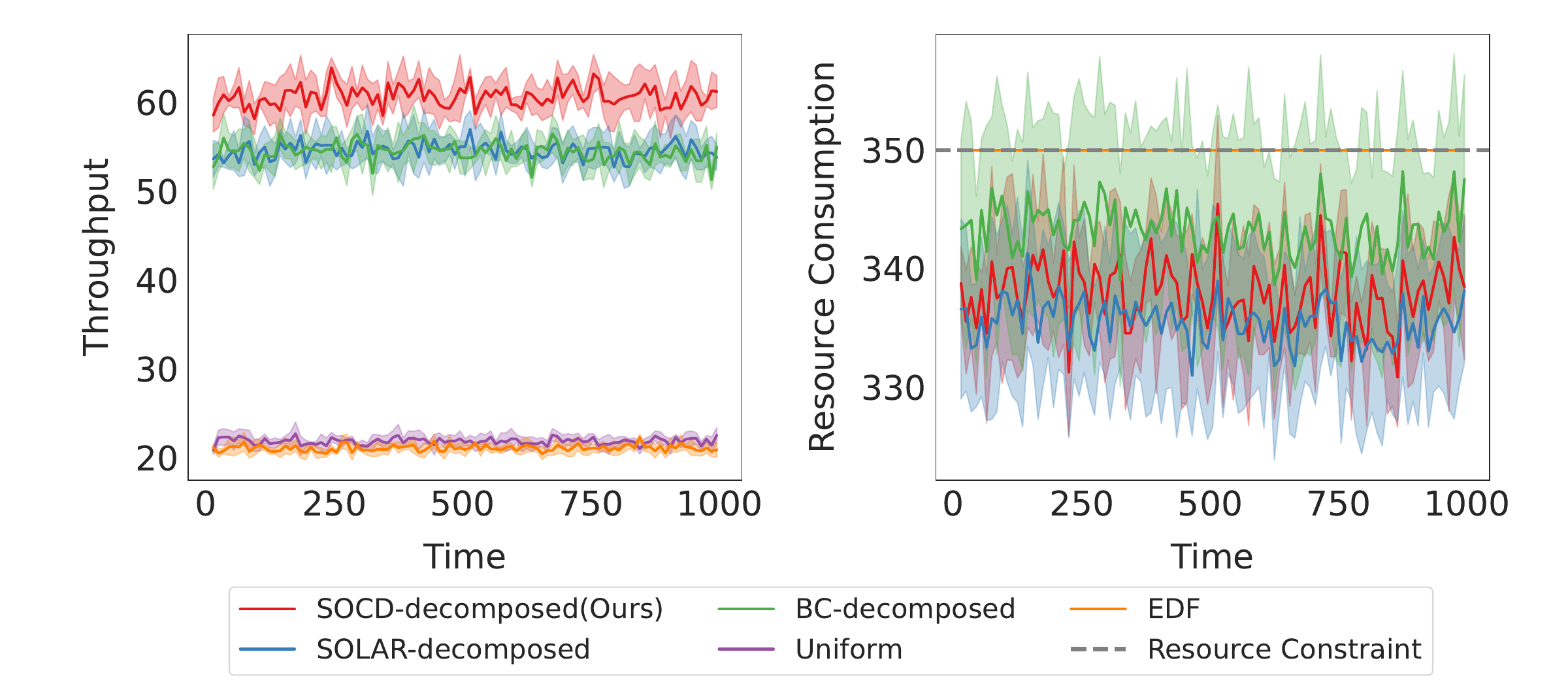}
	\caption{Comparison of various algorithms under an resource constraint of 350 as the number of users increases to 100. The left plot shows throughput, while the right plot illustrates resource consumption over time. All data are averaged across 10 time points.}
	\label{fig:100user}
\end{figure}

In the three environments characterized by different dynamic changes discussed above, it is clear that SOLAR's performance does not significantly exceed that of the behavior policy or the BC policy. In contrast, SOCD consistently demonstrates better generalization with more stable improvements, thanks to its diffusion-based policy component.

\subsubsection{Comparative Analysis}

\begin{figure}[htbp]
    \centering
    \includegraphics[width=1.0\linewidth]{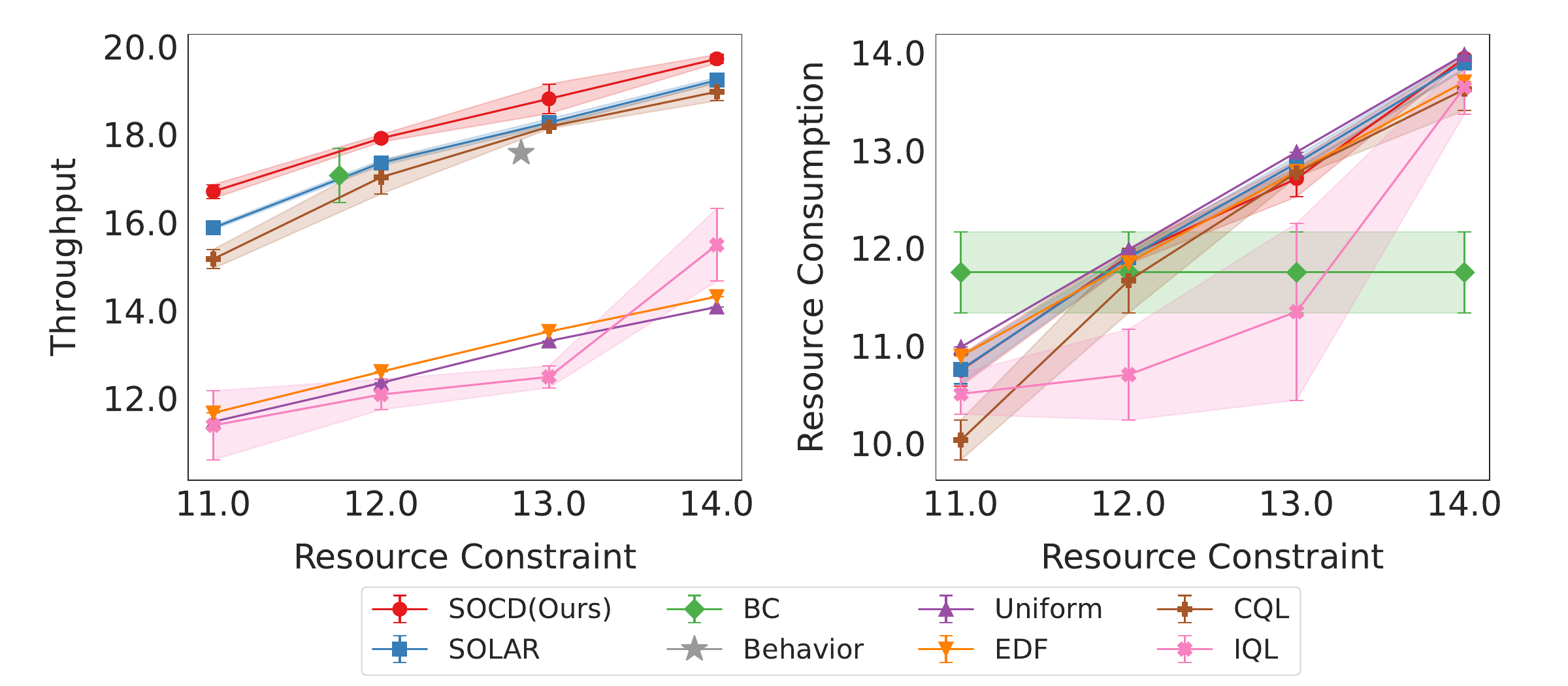}
    \caption{{Comparison of throughput and resource consumption among different baseline algorithms under varying resource constraints.}}
    \label{fig:ablation_study}
\end{figure}

{As illustrated in Figure~\ref{fig:ablation_study}, we evaluate the resource consumption of our proposed SOCD against various baselines under varying resource constraints. The results demonstrate that SOCD effectively learns an efficient policy, clearly outperforming traditional scheduling methods, such as Uniform and EDF. Furthermore, SOCD significantly outperforms all other RL-based algorithms, including BC, CQL \cite{kumar2020conservative}, and IQL \cite{kostrikov2021offlines}. This indicates that our DRL-based approach can successfully generalize beyond the suboptimal behavior policy embedded in the offline dataset to effectively utilize the offline data to learn improved, constraint-aware policies.}

{Notably, SOCD consistently outperforms SOLAR \cite{li2024offline}, the strongest RL baseline, in terms of resource consumption efficiency. When the resource constraint is tight (e.g., between 11.0 and 13.0), most baseline algorithms exhibit a steep increase in resource utilization, closely mirroring the constraint limits. In stark contrast, SOCD maintains a significantly lower and more stable resource consumption level. It intelligently conserves resources under stricter constraints and only increases utilization when the constraint is fully relaxed to 14.0, further highlighting the superiority of SOCD.}

\subsubsection{Discussion}

{Regarding estimation bias in offline resource estimation, SOCD mitigates potential extrapolation errors through its Diffusion-based Behavior Cloning backbone, which inherently anchors the policy to the behavior data support. This design limits the significant deviations that typically cause severe bias. Empirically, our results (Figures~\ref{fig:pois}--\ref{fig:ablation_study}) confirm that SOCD consistently respects actual resource constraints during online evaluation, demonstrating that the offline estimates are sufficiently accurate to guide the Lagrangian optimization.}

{Building upon this feasibility foundation, the core advantage of SOCD lies in its policy representation. As detailed in Section~\ref{subsec:baselines}, SOLAR serves as a representative baseline for offline RL using a standard Gaussian policy network within a constrained framework similar to ours. Therefore, the comparison between SOCD and other baseline DRL-based algorithm, such as CQL \cite{kumar2020conservative}, IQL \cite{kostrikov2021offlines}, and SOLAR \cite{li2024offline} effectively functions as an comparative analysis, isolating the contribution of the diffusion model against a Gaussian baseline. }

{In complex network scheduling, the distribution of optimal actions is inherently multi-modal due to the combinatorial nature of user demands and channel states. Unimodal Gaussian policies often struggle to approximate such complex distributions, leading to suboptimal resource utilization and instability. This limitation is clearly evidenced in our experiments: in high-dimensional and complex settings such as the 3-hop network (Figure~\ref{fig:3hop}) and the large-scale 100-user scenario (Figure~\ref{fig:100user}), the Gaussian-based SOLAR fails to efficiently utilize resources or match the throughput of the behavior policy. In contrast, SOCD leverages the high expressivity of the diffusion model to accurately capture these multi-modal distributions. This capability allows SOCD to achieve significantly higher throughput and stability, quantitatively validating that the adoption of a diffusion-based policy is a critical factor in solving complex delay-constrained scheduling problems.}

\section{Conclusion}\label{sec:con}
This paper presents \underline{S}cheduling by \underline{O}ffline Learning with \underline{C}ritic Guidance and \underline{D}iffusion Generation (SOCD), an offline deep reinforcement learning (DRL) algorithm designed to learn scheduling policies from offline data. {SOCD introduces an innovative diffusion-based policy network, augmented by a sampling-free critic network that provides policy guidance. By integrating Lagrangian multiplier optimization into the offline reinforcement learning framework, SOCD effectively learns high-quality constraint-aware policies, eliminating the need for online system interactions during training and relying solely on offline data to autonomously refine the policies.} Experiments on simulated datasets demonstrate SOCD's robustness to diverse system dynamics and its superior performance compared to existing methods.

{For future work, we identify several promising directions to further enrich this framework. First, the offline-trained SOCD policy can serve as a robust ``warm start'' for online fine-tuning. This would enable real-time adaptation and more precise updates to the Lagrange multiplier once environment interaction is permissible. Second, we aim to scale the framework to distributed edge computing systems, addressing more complex and heterogeneous constraints to broaden the practical impact of this work. Finally, to facilitate reproducibility and future research, we will release the complete source code and relevant datasets on a public repository.}

\section{Acknowledgement}

This work was supported by the National Natural Science Foundation of China Grant 52494974.

\bibliographystyle{IEEEtran}
\bibliography{reference}

\begin{IEEEbiography}[{\includegraphics[width=1in,height=1.25in,clip,keepaspectratio]{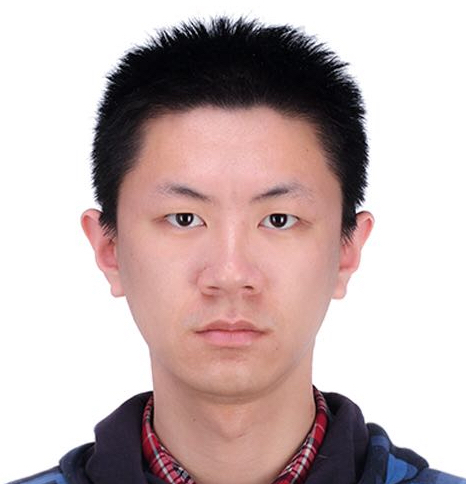}}]{Zhuoran Li}
received the B.E. degree in 2020
in electronic engineering from Tsinghua University,
Beijing, China, where he is currently working toward
the Ph.D. degree in the Institute for Interdisciplinary Information Sciences (IIIS) (headed by Prof. Andrew Yao), Tsinghua University, advised by Prof. Longbo Huang. His research interests include reinforcement learning, multi-agent systems, generative models and network optimization.
\end{IEEEbiography}
\vspace{-30pt}
\begin{IEEEbiography}[{\includegraphics[width=1in,height=1.25in,clip,keepaspectratio]{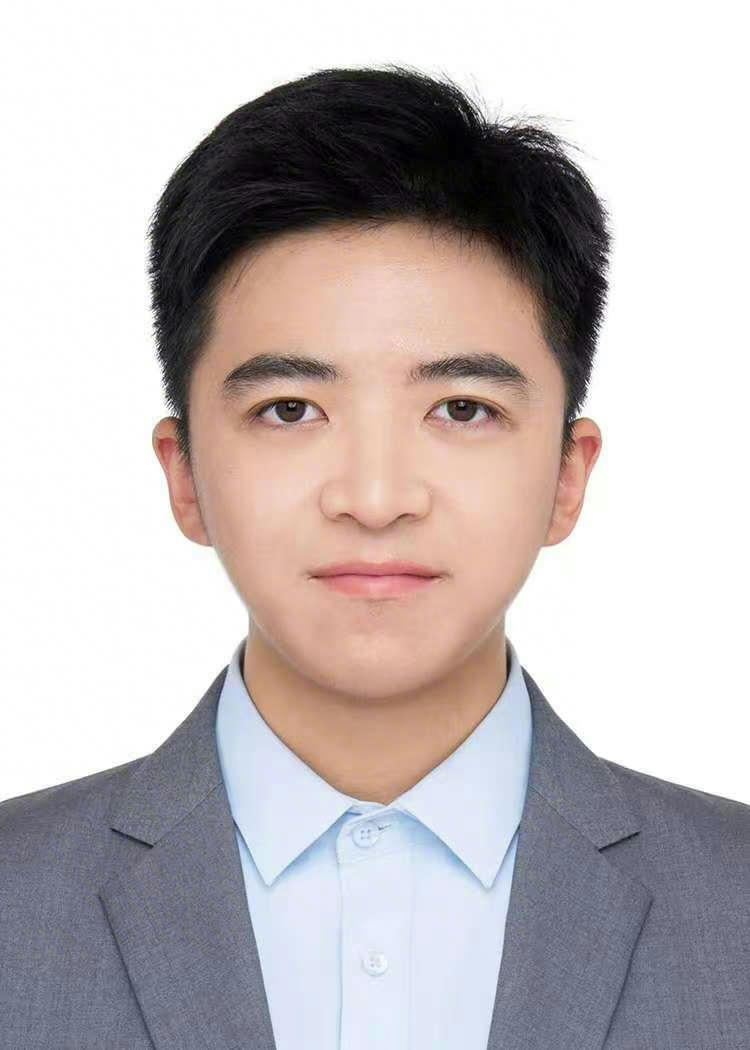}}]{Ruishuo Chen}
received the B.Sc. degree in Statistics from the School of Mathematics,
Nanjing University, Nanjing, China, in 2025. He is currently working toward
the M.Sc. degree at the Institute for Interdisciplinary Information Sciences
(IIIS), Tsinghua University, Beijing, China, under the supervision of
Prof. Longbo Huang. His research interests include reinforcement learning and large language models.
\end{IEEEbiography}
\begin{IEEEbiography}[{\includegraphics[width=1in,height=1.25in,clip,keepaspectratio]{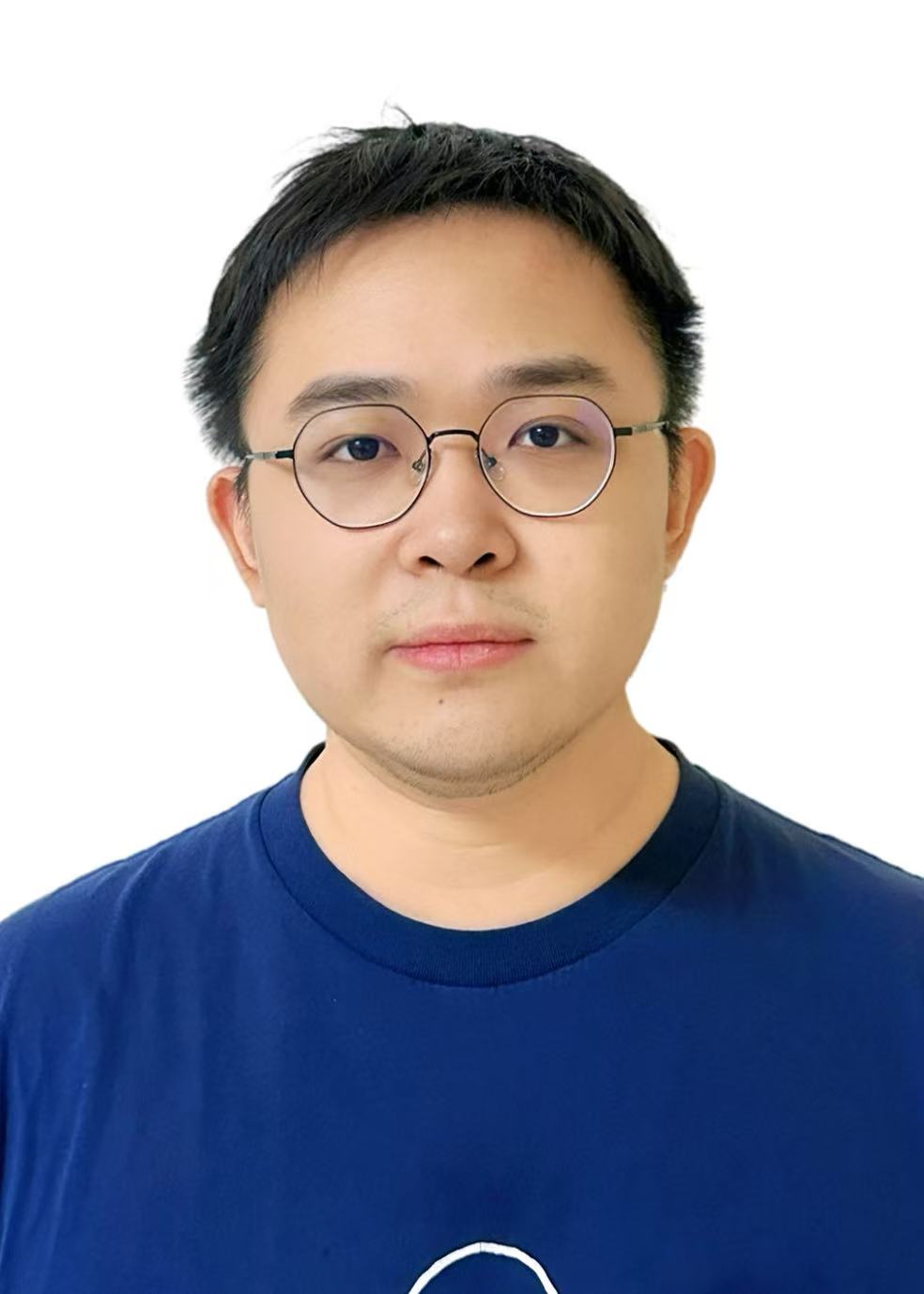}}]{Hai Zhong}
received the Master's degree in Electrical and Computer Engineering from the Georgia Institute of Technology, Atlanta, USA, in 2020. He is currently working toward the Ph.D. degree in the Institute for Interdisciplinary Information Sciences (IIIS), Tsinghua University, China. His research interests include reinforcement learning and multi-agent systems.
\end{IEEEbiography}
\vspace{-400pt}
\begin{IEEEbiography}[{\includegraphics[width=1in,height=1.25in,clip,keepaspectratio]{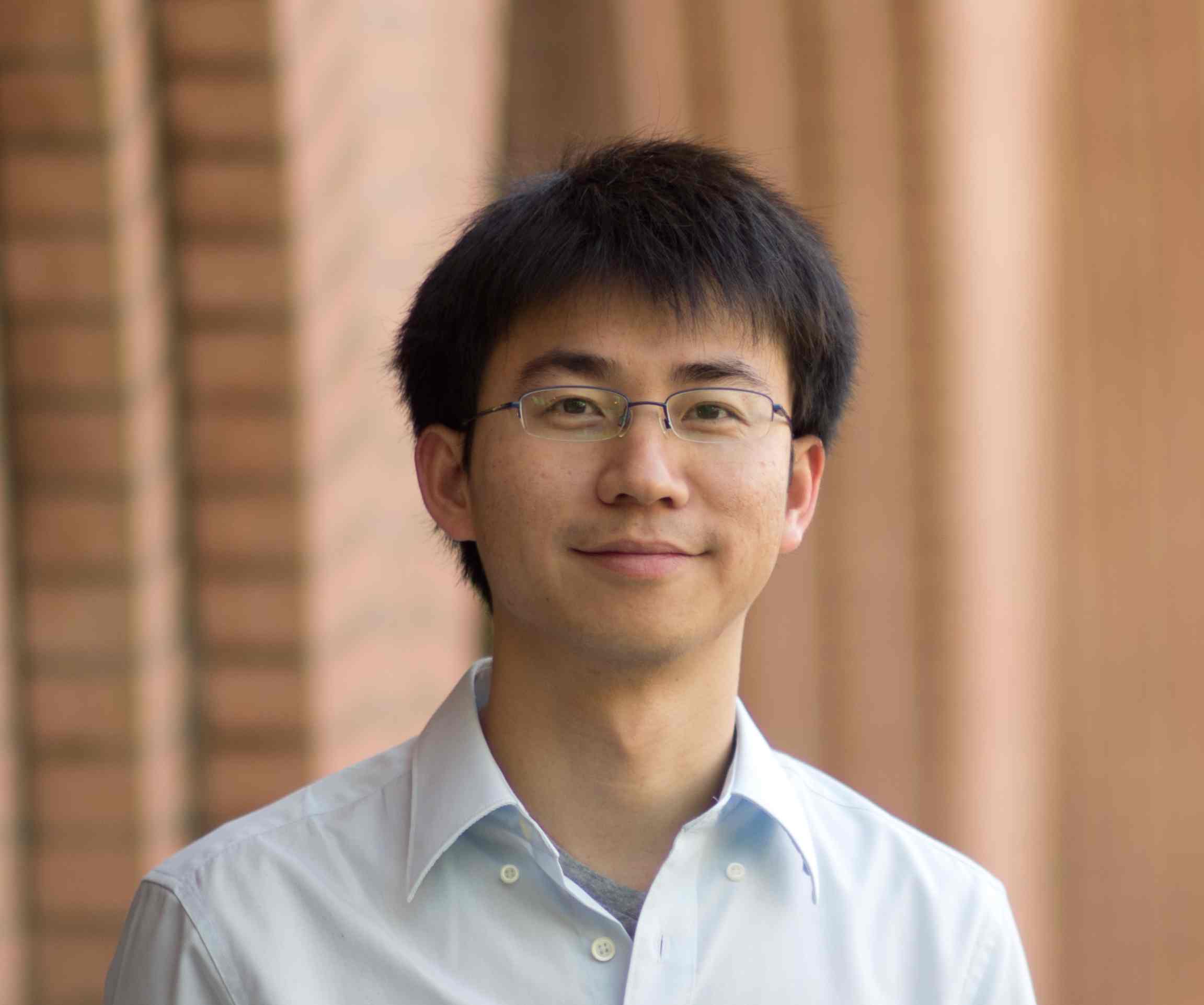}}]{Longbo Huang}
Prof. Longbo Huang is a full professor at the Institute for Interdisciplinary Information Sciences (IIIS) at Tsinghua University, Beijing, China. He has held visiting positions at the LIDS lab at MIT, the Chinese University of Hong Kong, Bell-labs France, and Microsoft Research Asia (MSRA). He was a visiting scientist at the Simons Institute for the Theory of Computing at UC Berkeley in Fall 2016. Prof. Huang serves/served as the General Chair for ACM Sigmetrics 2021, the TPC co-chair for ACM Sigmetrics 2027, IEEE WiOpt 2024, ITC 2022, IEEE WiOpt 2020 and NetEcon 2020. Prof. Huang serves/served on the editorial board for IEEE JSAC, IEEE TCOM, ACM ToMPECS,  IEEE/ACM TON, PEVA and IEEE TPAMI. He is an ACM Distinguished Member, CCF Distinguished Member, IEEE Senior Member, an ACM Distinguished Speaker and an IEEE ComSoc Distinguished Lecturer. Prof. Huang received the Outstanding Teaching Award from Tsinghua university in 2014, the Google Research Award and the Microsoft Research Asia Collaborative Research Award in 2014, and was selected into the MSRA StarTrack Program in 2015. Prof. Huang won the ACM SIGMETRICS Rising Star Research Award in 2018 and the ACM Sigmetrics 2025 Best Paper Award. 
\end{IEEEbiography}

\end{document}